\documentclass{article}

\usepackage{arxiv}

\usepackage[utf8]{inputenc} 
\usepackage[T1]{fontenc}    
\usepackage{hyperref}       
\usepackage{url}            
\usepackage{booktabs}       
\usepackage{amsfonts}       
\usepackage{nicefrac}       
\usepackage{microtype}      
\usepackage{lipsum}
\usepackage{graphicx}
\usepackage{booktabs}
\usepackage{algpseudocode}
\usepackage{multirow}  
\usepackage{amsfonts,amsmath,amsthm}
\usepackage{graphicx,subcaption,color,xcolor} 
\usepackage{hyperref,float} 
\usepackage{amsthm}

\theoremstyle{remark}
\newtheorem{remark}{Remark}
\graphicspath{ {./images/} }

\usepackage[linesnumbered,ruled,vlined]{algorithm2e}
\title{iPINNER: An Iterative Physics-Informed Neural Network with Ensemble Kalman Filter}

\author{
Binghang Lu \\
  School of Electrical and Computer Engineering\\
  Purdue University\\
  610 Purdue Mall, West Lafayette, 47907, IN \\
  \texttt{lu895@purdue.edu} \\
   \And
Changhong Mou \\
 Department of Mathematics\\
  Purdue University\\
  610 Purdue Mall, West Lafayette, 47907, IN\\
  \texttt{mouc@purdue.edu} \\
  \And
Guang~Lin \\
	Department of Mathematics and School of Mechanical Engineering \\
        Purdue University\\
	610 Purdue Mall, West Lafayette, 47907, IN \\
	\texttt{guanglin@purdue.edu} \\
}

\begin{document}
\maketitle
\begin{abstract}
 Physics-informed neural networks (PINNs) have emerged as a powerful tool for solving forward and inverse problems involving partial differential equations (PDEs) by incorporating physical laws into the training process. However, the performance of PINNs is often hindered in real-world scenarios involving noisy observational data and missing physics, particularly in inverse problems. In this work, we propose an iterative multi-objective PINN ensemble Kalman filter (iPINNER) framework that improves the robustness and accuracy of PINNs in both forward and inverse problems by using the \textit{ensemble Kalman filter} and the \textit{non-dominated sorting genetic algorithm} III (NSGA-III). Specifically, NSGA-III is used as a multi-objective optimizer that can generate various ensemble members of PINNs along the optimal Pareto front, while accounting the model uncertainty in the solution space. These ensemble members are then utilized within the EnKF to assimilate noisy observational data. The EnKF's analysis is subsequently used to refine the data loss component for retraining the PINNs, thereby iteratively updating their parameters. The iterative procedure generates improved solutions to the PDEs. The proposed method is tested on two benchmark problems: the one-dimensional viscous Burgers equation and the time-fractional mixed diffusion-wave equation (TFMDWE). The numerical results show it outperforms standard PINNs in handling noisy data and missing physics. 
\end{abstract}


\section{Introduction}
The rapid advancement of machine learning and artificial intelligence has profoundly influenced a wide range of scientific and engineering disciplines \cite{olier2021transformational,thiyagalingam2022scientific,karpatne2018machine,cuomo2022scientific,cai2021physics,guo2023learn,lin2025energy,qi2020using,kharazmi2021identifiability,zheng2022data,mou2023combining}. Among these advancements, \textit{physics-informed neural network} (PINN) \cite{cai2021physics,cuomo2022scientific,mao2020physics} has emerged as a powerful tool for solving complex partial differential equations (PDEs) by integrating physical laws directly into the learning process. PINN leverages the expressive capabilities of neural networks to approximate solutions to PDEs, offering a mesh-free and flexible alternative to traditional numerical methods. However, the effectiveness of PINN can be significantly impaired in the presence of noisy and sparse observational data (forward problem) \cite{cai2021physics,yang2021b,satyadharma2024assessing}, or missing physics, such as unknown coefficients in PDEs (inverse problem) \cite{zou2024correcting,jiang2021hybrid,chung2024hybrid,guo2021construct}, which are common challenges in real-world applications.

With perfect data, PINN is successfully used in forward and inverse problems. To train such a PINN, one needs to minimize a multi-objective loss function that includes the PDE residual (residual loss), initial conditions, boundary conditions (boundary loss), and data discrepancies (data loss) \cite{raissi2019physics,mao2020physics,shukla2020physics,lu2021physics,zhang2022analyses,rasht2022physics,xu2023transfer,zhou2023damageIMAC, lu2025evolutionary, zhang2025constrained}. They are widely used in many different problems.
For example, Mao et al. \cite{mao2020physics} utilized PINNs to infer density, velocity, and pressure fields for the one-dimensional Euler equations based on observed density gradient data. Similarly, Rasht et al. \cite{rasht2022physics} employed PINNs for full waveform inversions in seismic imaging to determine wave speed from observational data. Despite the success of PINNs, there are still several challenges. First, PINN uses the soft constraints which tend to minimize the sum of the PDE residual, boundary, and data losses with appropriate weights, however, the imbalance among different loss functions during training period may result in unusually expensive  training costs \cite{wong2022learning}. 
This can happen when certain terms dominate or vanish prematurely, resulting in inefficient training and therefore leading to non-optimal results. 
Second, PINNs are usually sensitive to available noisy data, which in fact is a very common setting in real-world applications. Indeed, noisy or imperfect data can ``mislead'' the training and therefore generate inaccurate results with inappropriate neural network parameters; the model errors in forward problem or only missing physics in inverse problem put the problem more challenging. 
Third, in inverse problems where only partial physics are known, i.e., PDEs with unknown parameters, the accuracy of traditional PINNs diminishes. This is because the PDE residual loss relies on both the neural network derivatives and the unknown parameters, which makes accurate inference challenging without balanced losses in training and access to high-quality data.


This paper proposes an integrated \textit{\textbf{iPINNER}} framework that combines 
the \textbf{I}terative \textbf{P}hysics \textbf{I}nformed \textbf{N}eural \textbf{N}etwork with \textit{\textbf{E}nsemble Kalman Filte\textbf{r}} (EnKF) \cite{Evensen_1996,evensen2003ensemble,KF_original}
to solve PDEs in both forward and inverse settings with noisy observational data. 
The iPINNER uses reference-point-based non-dominated sorting approach (hereby, referred to as NSGA-III) \cite{deb2000fast,deb2001controlled,deb2013evolutionary} to solve the multi-objective loss function in the original PINN.
Specifically, iPINNER employs NSGA-III to generate ensemble members of PINNs within the optimal Pareto front where these ensemble members are further used as forecast model results in EnKF, together with available observation data to iteratively refines the PINNs by updating its data loss function. 
The iPINNER framework integrates the advantages of two methods, \textit{evolution multi-objective optimizer}, i.e., NSGA-III and \textit{ensemble Kalman filter} (EnKF). The former can provide more balanced and effective training for PINNs with a multi-objective loss function while the optimal solutions from NSGA-III consist of various members on the optimal Pareto front, expressing the model uncertainty. 
In particular, NSGA-III treats each component of PINN loss as distinct objectives and in the training process, the non-dominated sorting and crowding distance calculation methods is employed \cite{lu2023nsga}. On the other hand, the latter, i.e., EnKF can be used to assimilate model and observational data to find the optimal solutions in the Bayesian sense.
While the original Kalman filter only handles linear systems \cite{KF_original}, the ensemble Kalman filter (EnKF) and its variant \cite{Anderson2001,Evensen_1996,evensen2003ensemble,mou2023efficient,chen2020predicting,popov2021multifidelity} extends it to a wider range of problems by using a Monte Carlo approach. Essentially, the EnKF begins with a probability distribution (represented by ensemble members in the forecast) and a likelihood function for observed data, then applies Bayes’ theorem to update this distribution (the “analysis” or posterior) once new observations are introduced. However, neural network–based PDE solvers often struggle to generate ensemble members that capture model uncertainty (model errors). The EnKF has
therefore, this iPINNER approach iterately refines the PINNs while leveraging the strengths of NSGA-III's multi-objective optimization capabilities and EnKF's denoising to improve the accuracy and robustness. 
{
        The proposed iPINNER framework is general and can be reformulated with different trial spaces once the PDE problem is recast as an optimization problem. In this work, we adopt a neural representation $u(x,t;\theta)$, where $\theta$ is the neural parameters, for the two reasons: (1) it can encode physical laws (e.g., PDE constraints, energy dissipation) and (2) it provides mesh-free automatic differentiation for residual evaluation.
}

This method can be used in both forward and inverse problem settings for the given PDEs. 
It is also important to note that, in the inverse problem setting for PINNs, there are two different approaches:
(I) The unknown physical parameters are treated as additional independent variables and included as inputs to the neural network, which is trained over a range of parameter values \cite{chen2023reduced,chen2022autodifferentiable};
(II) The physical parameters are treated as trainable variables. While they do not explicitly appear in the network architecture, they affect the training process through their contribution to the loss function via the PDE residual. In this paper, we use the second approach for the inverse problem, i.e., putting unknown physics term as the trainable variables which directly contributes to the PDE residual in PINN's loss function.  
In brief summary, our primary contribution includes the following:

\begin{enumerate}
    \item We employ the NSGA-III algorithm to treat each component of the PINN loss as an individual objective and use Non-dominated Sorting (NDS) and gradient decent method to optimize these objectives. Numerical tests show that this multi-objective approach helps the PINN avoid local minima and better satisfy physical and data constraints.
    \item We proposed the novel iPINNER framework that integrates Ensemble Kalman Filter (EnKF) with the NSGA-III-optimized PINN ensemble. The EnKF utilizes solutions from the Pareto front (as predictions) and assimilates observational data to update the state variables, which in turn refines the PINN training process. Experiments on both forward and inverse PDE/FPDE problems show that the framework significantly improves prediction accuracy in the presence of model imperfections and noisy data.
    \item We demonstrate that iPINNER significantly improves prediction accuracy over traditional PINNs in scenarios with incomplete physical knowledge and noisy observations. While conventional PINNs often fail under these conditions due to loss imbalance and sensitivity to noise, our framework leverages filtered observational data via EnKF to guide the model toward the ground truth.
    \item iPINNER also achieves strong performance in inverse PDE and FPDE problems, effectively recovering missing physical information by combining ensemble-based data assimilation with multi-objective training.
\end{enumerate}

Compared to methods that rely solely on multi-objective optimization, the iPINNER framework shows substantial improvements when only noisy data are available. The proposed framework introduces a novel iterative scheme that integrates multi-objective optimization with the Kalman filter to identify optimal solutions under noise-dominated conditions—an approach not previously explored in the PINN literature. The rest of the paper is organized as follows. In Section \ref{sec:general}, we introduce the proposed integrated model that combines NSGA-III optimizer (Section \ref{sec:nsga}) and ensemble Kalman filter (Section \ref{ss:enkf}) in the PINN framework (Section \ref{sec:pinn}). Section \ref{sec:numeric} presents numerical test results for both forward and inverse problems for the proposed framework. Specifically, we test two different problems: (1). one-dimensional viscous Burgers equation in Section \ref{sec:burgers}, and  (2). time-fractional mixed diffusion-wave equations (TFMDWEs) in Section \ref{sec:frac}. Finally, conclusions and future research directions are discussed in Section \ref{sec:conclusion}.

\section{General framework \label{sec:general}}

\begin{figure}[H]

    \centering
    \includegraphics[width=1.1\linewidth]{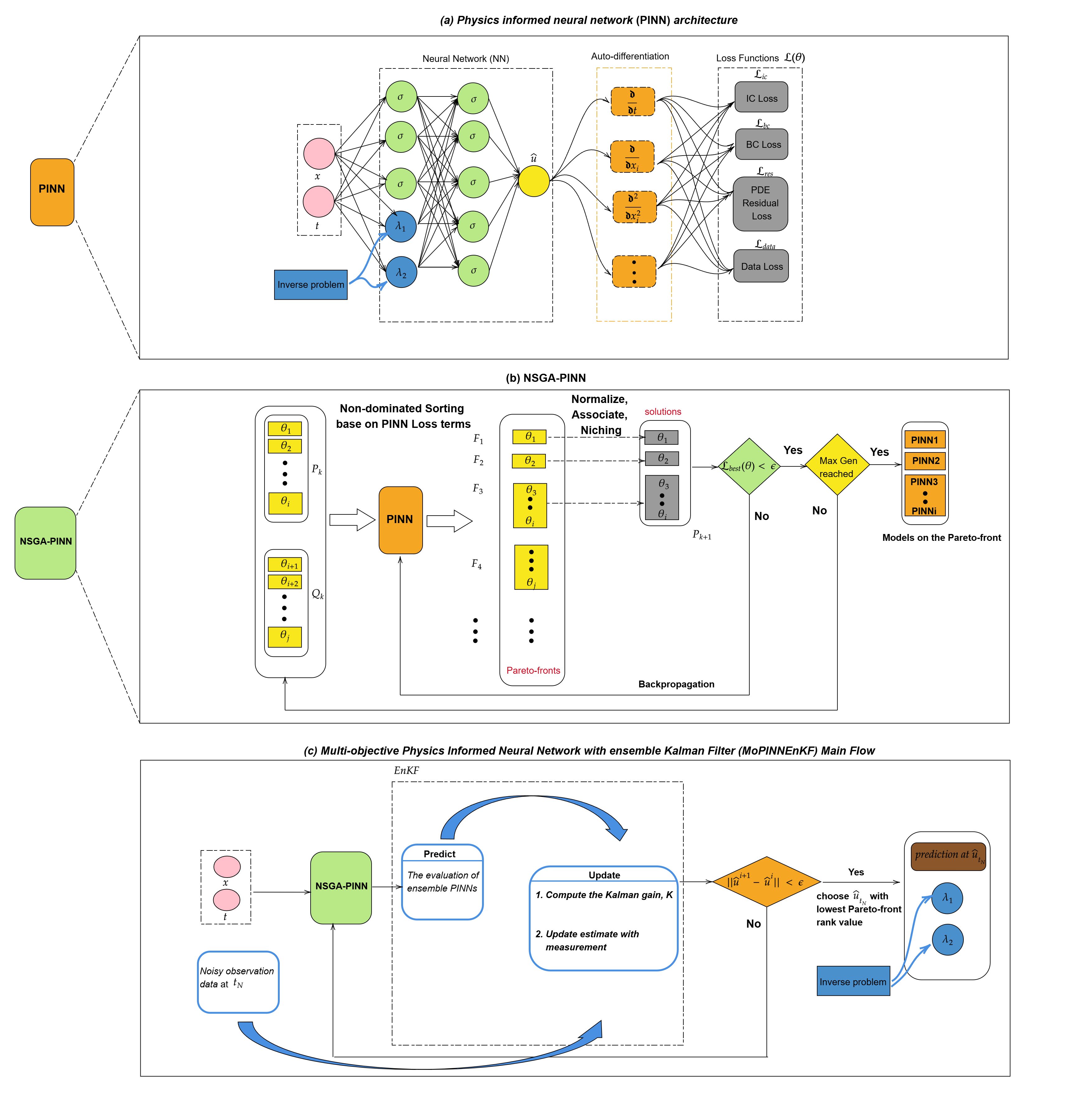}
 \caption{%
 Schematic of the proposed workflow for iPINNER framework for forward and inverse problem (parameter estimation or missing physics). (a) Firstly, the neural network takes spatiotemporal inputs $(x,t)$ and, through automatic differentiation, enforces partial differential equation (PDE) constraints, boundary and initial conditions (ICs/BCs), as well as noisy observational data via loss functions. (b) Secondly. a multi‐objective
loss and multiple candidate networks are then refined using a multi‐objective optimizer (NSGA-III), yielding robust solutions consistent with both data and governing physics. (c) The main flow of iPINNER that uses ensemble output from NSGA-PINN as initial observation and combine 
    }   
     \label{fig:illustration}
\end{figure}

In this section, we introduce the physics‐informed neural network (PINN), the Kalman filter, and the Non-Dominated Sorting Genetic Algorithm-III (NSGA-III), and then describe the framework that integrates these components as illustrated in Figure~\ref{fig:illustration} and Algorithm~\ref{alg:inferrring}. 
The framework employs PINNs to integrate potentially incomplete physical information through partial differential equation (PDE)‐based loss functions, while the Kalman filter assimilates observational data in real time to optimally estimate state variables with PINN models.
Subsequently, NSGA-III is employed to optimize the PINN ensemble by treating various loss components (e.g., PDE residual, data mismatch) as separate objectives, efficiently exploring the parameter space to identify a set of non-dominated solutions (the Pareto front). This integrated approach ensures that each component informs and improves the others, leading to a comprehensive, data‐driven methodology that remains faithful to the underlying physics.

\subsection{Physics Informed Neural Network (PINN)\label{sec:pinn}}
To illustrate the framework of physics informed neural networks (PINNs), we start with the general nonlinear PDE which takes the form \cite{lu2021physics,karniadakis2021physics,mao2020physics}:
\begin{eqnarray}\label{eq:PDE}
u_t + \mathcal{N}[u] = 0,\ x \in \Omega, \ t\in[0,T],
\end{eqnarray}
with a suitable initial condition and Dirichlet boundary conditions, where $u(t,x)$ denotes the latent (hidden) solution, $\mathcal{N}[\cdot]$ is a nonlinear differential operator, and $\Omega$ is a subset of $\mathbb{R}^D$. 
We define the PDE residual as a function $f$:
\begin{eqnarray}\label{eq:residual}
f = 
u_t + \mathcal{N}[u].
\end{eqnarray}
PINN framework finds a neural network (NN), which is
parametrized by a set of parameters $\theta$, i.e., $ \hat{u}_\theta(z)$  to approximate the solution to the PDE. 
To determine the parameter set $\theta$ that defines the model, PINN solves a optimization problem minimizing a suitably constructed loss function, which incorporates contributions 
from the differential equation \(\mathcal{L}_\mathcal{F}\), the boundary conditions 
\(\mathcal{L}_\mathcal{B}\), and any available data \(\mathcal{L}_\mathrm{data}\), 
with each component appropriately weighted.
Specifically, the optimization problem yields the following:
\begin{align}\label{eq:op}
      \theta^* \;=\; \arg\min_{\theta} \Bigl( 
   \omega_{ic}\,\mathcal{L}_{ic}(\theta)
    \;+\; 
    \omega_{bc}\,\mathcal{L}_{bc}(\theta)
     \;+\; 
     \omega_{res}\,\mathcal{L}_{res}(\theta)
    \;+\; 
     \omega_{data}\,\mathcal{L}_\text{data}(\theta)
  \Bigr).
\end{align}
where 
\begin{eqnarray}\label{eq:loss_bcic}
\mathcal{L}_{ic} = \frac{1}{N_{ic}} \sum_{i=1}^{N_{ic}} \left( u(x_i, t_i) - u_i \right)^2,
\end{eqnarray}
\begin{eqnarray}\label{eq:loss_bcic}
\mathcal{L}_{bc} = \frac{1}{N_{bc}} \sum_{i=1}^{N_{bc}} \left( u(x_i, t_i) - u_i \right)^2,
\end{eqnarray}
\begin{eqnarray}\label{eq:loss_pde}
\mathcal{L}_{res} = \frac{1}{N_{res}} \sum_{j=1}^{N_{res}}\left( 
u_t + \mathcal{N}[u] 
\right)^2 \bigg|_{(x_j, t_j)}.
\end{eqnarray}
and

\begin{eqnarray}\label{eq:loss_data}
\mathcal{L}_\text{data} = \frac{1}{N_\text{data}} \sum_{k=1}^{N_\text{data}} \left( u(x^d_k, t^d_k) - u^d_k \right)^2.
\end{eqnarray}
Here $\{(x_i, t_i)\}$ are the number of points sampled at the initial/boundary locations 
and in the entire domain, respectively, and $\{u^d_k\}$ is a set of $u$ values that are accessible at $\{(x^d_i, t^d_i)\}$; $w_{res}$, $w_{ic}$, $w_{bc}$, and $w_{\text{data}}$ are the weights used to balance the interplay among the four loss terms.
These weights affects both the convergence rates and the obtained minimum for the optimization problem~\eqref{eq:op} \cite{psaros2022meta,bai2023physics}, therefore it is crucial to choose their values. 
In \cite{lu2023nsga}, the authors integrate the NSGA-III algorithm into the optimization framework for PINNs, and we use the same strategy in this paper.
%

\subsection{Non‐Dominated Sorting Genetic Algorithm-III (NSGA-III) \label{sec:nsga}}
Multi-objective optimization problems frequently arise in various scientific and engineering fields, including artificial intelligence \cite{tian2021evolutionary}, data mining \cite{hong2021evolutionary}, software engineering \cite{li2019quality}, scheduling \cite{jozefowiez2008multi}, bioinformatics \cite{rockendorf2022design}, and economics \cite{park2024techno}, many of which involve high-dimensional spaces. 
In what follows, we will introduce the nondominated sorting genetic algorithm (NSGA-III) for multi-objective optimization problems and how it is used to reduce \textit{uncertainty} in PINN frameworks. 

\subsubsection{Multi-objective Optimization\label{sec：nsga}}
The multi-objective optimization problem is defined as follows:
Given $m\in \mathbf{N}$, the \emph{$m$-objective function} is defined as $f(x)=(f_1(x),\ldots, f_m(x))$ where $x\in\Omega$, and $f_i\colon \Omega\rightarrow \mathbf{R}$ for a given search space~$\Omega$.
Other than in single-objective optimization, there is usually no solution that minimizes all $m$ objective functions simultaneously.
Suppose that there are two solutions $x,y$,  $x$ \emph{dominates} $y$ referring to $x\preceq y$ if and only if $f_j(x)\le f_j(y)$ for all $1\le j\le m$.
If there exists a $j_0$ such that $f_{j_0}(x)<f_{j_0}(y)$, we refer that $x$ \emph{strictly dominates} $y$, denoted by $x\prec y$.
A solution is defined as \emph{Pareto-optimal} if it is not strictly dominated by any other solution.
Here, the set of objective values of Pareto-optimal solutions is denoted as \emph{Pareto front}.

To solve such optimization problems, one can consider population-based evolutionary algorithms, which are usually more effective than traditional mathematical programming in finding a set of solutions that balance different objective functions. 
However, it is becoming increasingly clear that expecting a single population-based optimization method to achieve convergence near the Pareto-optimal front and maintain a uniform distribution across the entire front in high-dimensional problems is impractical \cite{deb2011multi,zitzler2000comparison,goldberg1989messy}.
One remedy to tackle such an issue is to incorporate external mechanisms to support diversity maintenance, which could also alleviate the computational burden. Instead of exhaustively exploring the entire search space for Pareto-optimal solutions, the algorithm can initiate multiple predefined, targeted searches. 
The nondominated sorting genetic algorithm (NSGA-III), proposed by Deb and Jain \cite{deb2013evolutionary}, is based on this principle and has been shown to be an effective multi-objective optimization method within the evolutionary optimization framework. 
NSGA-III modifies NSGA-II \cite{deb2002fast} by incorporating a reference-point–based mechanism for many-objective problems, emphasizing population members that are non-dominated yet lie close to a set of specified reference points. 
NSGA-III initializes with a random population of size $N$.
In each iteration, the user generates an offspring population of size $N$ with mutation and/or crossover operators.
With a fixed population size, out of this total of $2N$ individuals, NSGA-III selects $N$ for the next iteration.
Because non-dominated solutions are preferred, the following ranking scheme establishes the dominance relation as the principal criterion for individual survival.
Individuals that are not strictly dominated by any other in the population are assigned rank $1$. 
Subsequent ranks are determined recursively: each unranked individual that is strictly dominated only by those with ranks $1,\ldots,k-1$ is assigned rank $k$. 
Intuitively, the lower an individual's rank, the more interesting it is.
Denote $F_i$ be the set of individuals with rank $i$, and let $i^*$ be the smallest integer such that
\begin{align}
\sum_{i=1}^{i^*} |F_i| \ge N.
\end{align}
All individuals with rank at most $i^* - 1$ retain to the next generation. In addition, $0 < k \le N$
individuals of rank $i^*$ must be selected so that the new population remains $N$, allowing the next iteration to proceed.

\subsubsection{Uncertainty}
In \cite{lu2023nsga}, the authors use the non-dominated sorting genetic algorithm (NSGA) to improve traditional stochastic gradient optimization methods (e.g., ADAM), enabling them to escape local minima more effectively.
Indeed, when complete data or fully known physics information is available, PINNs combined with NSGA can effectively solve PDEs in both forward and inverse problem settings \cite{moya2025conformalized}. However, when data availability is limited or the data are noisy, the framework may inadvertently incorporate uncertainties originating from noisy data in forward problems. Moreover, incomplete or missing physics information can lead to wrong solutions in inverse Problems.
On the other hand, NSGA-III provides an alternative method for quantifying uncertainty arising from data-driven loss functions and potentially incomplete physical models. Specifically, NSGA-III generates a set of Pareto-optimal solutions (see Section~\ref{sec：nsga}), which can be viewed as multiple realizations or candidate parameter configurations within the PINN framework. These solutions can be integrated into an ensemble Kalman filter (in Section~\ref{ss:enkf}), together with observational data, to approximate the posterior probability distribution. This approximation can then be iteratively utilized back to refine and correct the data loss in the PINN framework.
The NSGA-III algorithm used in PINN is outlined in Algorithm~\ref{alg:nsga-iii}. 

\begin{algorithm}[H]
\DontPrintSemicolon
\SetAlgoLined
\caption{NSGA-III in PINN's multi-objective optimization \label{alg:nsga-iii}}
\KwSty{Initialize a population $P_0$ consisting of $N$ individuals chosen independently and uniformly:}
$$
P_0 = \{\theta_0^{(i)}\}_{i=1}^N
$$
\KwSty{where $\theta_0^{(i)}$ denotes
the parameters of the $i^{th}$ neural network individual.}

\For{$t = 0, 1, 2, \ldots$}{
  Generate offspring population $Q_t$ of size $N$ using selection, crossover, and mutation operations.\;

  Set combined population $R_t \gets P_t \cup Q_t$.\;

  Apply \textit{fast non-dominated sorting} \cite{deb2002fast} to partition $R_t$ into non-dominated fronts $\mathcal{L}_1, \mathcal{L}_2, \dots$ based on the multi-objective loss function defined in \eqref{eq:op}.\;

  Identify the front index $i^* \ge 1$ such that:
  \[
    \sum_{i=1}^{i^*-1} |\mathcal{L}_i| < N \quad \text{and} \quad \sum_{i=1}^{i^*} |\mathcal{L}_i| \ge N
  \]\;

  Set $Z_t \gets \displaystyle \bigcup_{i=1}^{i^*-1} \mathcal{L}_i$.\;

  Select subset $\tilde{\mathcal{L}}_{i^*} \subseteq \mathcal{L}_{i^*}$ satisfying:
  \begin{align*}
    &\bigl|Z_t \cup \tilde{\mathcal{L}}_{i^*}\bigr| = N \\
    &\text{Selection based on reference points $R$ when maximizing the function $\mathcal{L}$.}      
  \end{align*}\;

  Update the population:
  \[
    P_{t+1} \gets Z_t \cup \tilde{\mathcal{L}}_{i^*}
  \]\;
}
\end{algorithm}

\subsection{Ensemble Kalman Filter (EnKF) \label{ss:enkf}}
Since the ensemble Kalman filter (EnKF) was first developed in geophysics \cite{evensen2003ensemble,Evensen_1996}, it has been widely used in atmospheric science, oceanography, and climate modeling. The EnKF involves two steps: (1). forecast (prediction) and (2). analysis (filtering). 
The forecast step is usually temporal propagation of state-space models:
\begin{align}
    x_{t} = F(x_{t-1})
\end{align}
In the update step, we assume that each observation is a linear combination of the state, perturbed by Gaussian noise. Formally, for a given state \(x_t\), the observation \(Y_t\) follows a normal distribution:
\begin{align}
Y_t \mid x_t \sim \mathcal{N}\bigl(H x_t,\ \sigma^o\bigr),\label{eq:noise}    
\end{align}
where $H$ is the obervational operator and $\sigma^o$ denotes the amplitude of the observation noise.
If the prediction distribution is normal,
\begin{align}
      \pi_{t\mid t-1} \;=\; \mathcal{N}\bigl(m_{t\mid t-1},\ P_{t\mid t-1}\bigr),
\end{align}
then the filter distribution is also normal,
\begin{align}
  \pi_t \;=\; \mathcal{N}\bigl(m_t,\ P_t\bigr),    
\end{align}
with
\begin{align} 
  m_t \;=\; m_{t\mid t-1} \;+\; K_t\bigl(y_t - H\,m_{t\mid t-1}\bigr),
  \quad
  P_t \;=\; \bigl(I \;-\; K_t\,H\bigr)\,P_{t\mid t-1},
\end{align}
where
\begin{align}   
  K_t \;=\; K\!\bigl(P_{t\mid t-1},\,R\bigr)
        \;=\; P_{t\mid t-1}\,H^\top\bigl(H\,P_{t\mid t-1}\,H^\top + R\bigr)^{-1}.
\end{align}
is the Kalman gain.
In the ensemble Kalman filter (EnKF), both $\pi_{t \mid t-1}$ and $\pi_t$ are approximated by equally weighted ensembles $\{x_t^{(i)}\}$ and $\{\widetilde{x}_t^{(i)}\}$. During the update step, one first employs the prediction sample to estimate $m_{t\mid t-1}$ and $P_{t\mid t-1}$. The filter sample is then constructed by transforming the prediction sample so that its mean and covariance align with the update equations. There are several possible approaches to carry out this transformation \cite{evensen2003ensemble,evensen2022}. In this paper, we use the plain ensemble Kalman filter (EnKF) and the forecast step, the prior is generated using the PINNs from NSGA-III algorithm. 
\subsection{A brief summary of the framework}
Algorithm~\ref{alg:inferrring} summarize the framework that infers model on the fly using generic-PINN-based ensemble Kalman filter. In brief, the algorithm integrates physics-informed neural networks (PINNs) with NSGA-III and the ensemble Kalman filter (EnKF) to iteratively infer models from observational data. Initially, the PDE solution is approximated by neural networks, whose parameters are optimized using gradient descent coupled with NSGA-III to produce a few PINNs with different neural network parameters which lies in the optimal Pareto front. These networks are then evaluated at observational points, and the ensemble Kalman filter updates the solution based on observed data, resulting in a posterior estimate. This posterior is then used to refine the PINN loss function, iteratively improving the model until convergence is reached based on a prescribed threshold.

\begin{algorithm}[H]
\caption{Algorithm for Inferring model on the fly using MoPINNEnKF \label{alg:inferrring}}
\label{alg:ensemble_pinn}
\begin{algorithmic}[1]
\State \emph{Represent the PDE solution by a neural network.}

\State \emph{Formulate the weighted loss function according to the PDE system:}
  \begin{eqnarray*}
    \mathcal{L} \;=\;
    \Bigl( 
      \omega_\mathcal{F}\,\mathcal{L}_\mathcal{F}(\theta)
      \;+\; \omega_\mathcal{B}\,\mathcal{L}_\mathcal{B}(\theta)
      \;+\; \omega_d\,\mathcal{L}_\text{data}(\theta)
    \Bigr).
  \end{eqnarray*}

\State \emph{Use $S$ steps of a gradient descent algorithm together with NSGA-III to update the parameters $\theta$, obtaining a cluster of PINNs:} 
\[
   \{\,u_{\theta_l}^{(1)}: \; l=1,\dots,N_s\},
\]
\emph{where $N_s$ is the number of offsprings in NSGA-III.}

\State \emph{Evaluate the PINNs at observational points $(x_k,t_k)$, $k=1,\cdots, N_{\mathrm{obs}}$:}
  \[
   u_{\theta_l}^{(1)}(x_1,t_1), \; u_{\theta_l}^{(1)}(x_2,t_2), \;\dots,\; u_{\theta_l}^{(1)}(x_{N_{\mathrm{obs}}},t_{N_{\mathrm{obs}}}).
  \]
  \emph{Here, $u_{\theta_l}^{(1)}(x_1,t_1)$ is interpreted in an ensemble sense.}

\State  \emph{Use the ensemble Kalman filter (EnKF) and the observational data set to obtain the posterior estimate of the data set:}
  \[
    \mathcal{D}^{(1)}(u)
    =
    \bigl\{\widetilde{u}_\theta^{(1)}(x_1,t_1),\;\widetilde{u}_\theta^{(1)}(x_2,t_2),\;\dots,\;\widetilde{u}_\theta^{(1)}(x_{N_{\mathrm{obs}}},t_{N_{\mathrm{obs}}}) \bigr\}.
  \]

\State  \emph{Use the data set $\mathcal{D}^{(1)}(u)$ to update the loss function in PINN:}
  \begin{eqnarray}\label{eq:loss_data-iter-1}
    \mathcal{L}_{\mathrm{o}} \;=\; \frac{1}{N_{\mathrm{obs}}} \sum_{k=1}^{N_{\mathrm{obs}}} 
    \bigl(\widetilde{u}_\theta^{(1)}(x_k, t_k) \;-\; u_k \bigr)^2.
  \end{eqnarray}

\State \emph{Repeat from Step 2 until}
  \[
    \frac{1}{N_{\mathrm{obs}}}\sum_{k=1}^{N_{\mathrm{obs}}} 
    \bigl\lvert u_\theta^{(m+1)}(x_k,t_k) \;-\; u_\theta^{(m)}(x_k,t_k)\bigr\rvert^2 
    \;<\;\epsilon_{\mathrm{iter}},
  \]
  \emph{where $\epsilon_{\mathrm{iter}}$ is a prescribed threshold.}

\end{algorithmic}
\end{algorithm}




\section{Numerical Results \label{sec:numeric}}
In this section, we test the proposed framework, hereafter referred to as iPINNER, using two different benchmark problems: (1) the one-dimensional viscous Burgers equation and (2) the one-dimensional time-fractional mixed diffusion-wave equations (TFMDWEs). For each of these problems, we consider two scenarios: (i) the forward problem and (ii) the inverse problem. Specifically, in the inverse problem setting, we assume the diffusion coefficient is unknown for the viscous Burgers equation, while the fractional order is unknown for the TFMDWEs.
Also, the observational data, $\{u^{obs}(x,t)\}$ are generated by introducing observational noise at sparse grid point, with a noise level of $\eta$. Specifically, this observational noise is drawn from a Gaussian distribution with zero mean, and its standard deviation is set equal to $\eta$ of the standard deviation of the true solution, $std(u^{truth}(x,t))$, at grid $x$. The mathematical formulation yields the following:
\begin{align}
    u^{obs}(x,t) = u(x,t) +\eta\,\cdot \mathcal{N}(0,std(u^{truth})).
\end{align}
To investigate the impact of noise levels on solutions of both forward and inverse problems, we consider three different values of the parameter $\eta$, corresponding to scenarios of small, medium, and large observational uncertainties, respectively:
\begin{enumerate}
    \item $\eta = 20\%$, representing observations with small uncertainty;
    \item $\eta = 50\%$, representing observations with medium uncertainty;
    \item $\eta = 80\%$, representing observations with large uncertainty.
\end{enumerate}
In addition, we compare our proposed \textit{iPINNER} method with two other approaches: (a) a PINN employing the traditional Adam optimizer, denoted as \textit{ADAM-PINN}, and (b) a PINN employing only the NSGA-III optimizer, denoted as \textit{NSGA-III-PINN}.

\subsection{Burgers Equation \label{sec:burgers}}
The one-dimensional viscous Burgers equation is a nonlinear partial differential equation frequently used as a benchmark \cite{lu2021deepxde,lu2021physics,raissi2019physics}. {The equation with Dirichlet boundary conditions is defined on the spatial} domain $\Omega = [-1,1]$ and temporal domain $[0,T]$ given by the following:
\begin{align}
&\frac{\partial u}{\partial t} - \nu \frac{\partial^2 u}{\partial x^2} + u \frac{\partial u}{\partial x} = 0,\quad x \in \Omega,\; t \in [0,T],\label{eq:burgers}\\[6pt]
&u(x,t) = 0,\quad \forall x \in \partial\Omega,\\[6pt]
&u(x,0) = -\sin(\pi x).
\end{align}
Here, $u(x,t)$ represents the solution over space and time, and $\nu$ is the viscosity chosen to be $0.01/\pi$. 
\subsubsection{PINN's Settings \label{sec:nn-setting}}
\begin{figure}[H]
    \centering
    \resizebox{\textwidth}{!}{  
    \begin{minipage}{\textwidth}
        \centering    \begin{subfigure}[b]{0.3\linewidth}
    \centering
        \includegraphics[width=\linewidth]{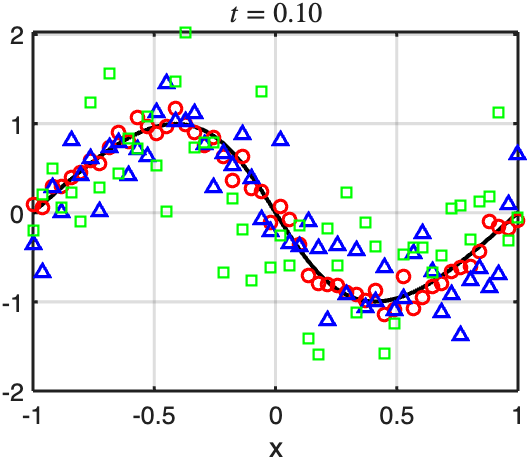}
        \caption{}
    \end{subfigure}
    \begin{subfigure}[b]{0.3\linewidth}
    \centering
        \includegraphics[width=\linewidth]{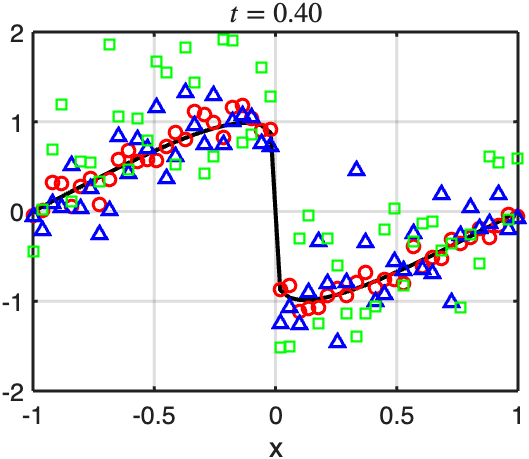}
        \caption{}
    \end{subfigure}
    \begin{subfigure}[b]{0.31\linewidth}
    \centering
        \includegraphics[width=\linewidth]{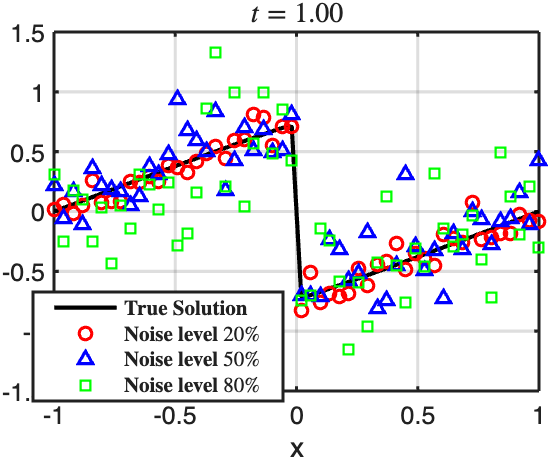}
        \caption{}
    \end{subfigure}
        \end{minipage}
    }
    \caption{
Observation data with different noise levels at time instances (a)~$t = 0.1$, (b)~$t = 0.4$, and (c)~$t = 1$ for the one-dimensional Burgers equation.
    }
    \label{fig:burgers-noise}
\end{figure}

\paragraph{Observational data}
To generate observational data used in PINN framework, we choose $100$ data points evenly for the initial and boundary conditions and $10^4$ interior points.
Figure~\textcolor{blue}{\ref{fig:burgers-noise}} shows the different noise level observation at different time instance for the one-dimensional viscous Burgers equation.
\paragraph{Neural network architecture} 
In order to make a fair comparison, with different optimizers, the PINN employs a consistent deep neural network architecture comprising 8 layers with 20 neurons in each layer.
During the training phase, models corresponding to each optimizer were collected until the training loss converged below a prescribed threshold $\epsilon$.
Specifically, the ADAM-PINN requires $5000$ epochs to achieve convergence, whereas the NSGA-III-PINN arrives at convergence within $4$ generations. The proposed iPINNER requires of $3$ generations with $1000$ epochs to ensure the convergence of training loss.

{
\paragraph{Training and Testing Data} The training data for the Burgers equation consist of three main components: Initial Condition (IC) points, Boundary Condition (BC) points, and Collocation Points. The IC points are sampled in space at the initial time $t = 0$, while the BC points are sampled in time along the spatial boundaries ($x = -1$ and $x = 1$). The collocation points are randomly selected from the interior of the spatiotemporal domain, with 100 points used in this study. The testing data are sampled over the domain $\Omega \times [0,T]$ with $\Omega = [0,1]$.  The spatial mesh size is of $\Delta x = 1/100$ and a temporal step size is $\Delta t = 0.01$.

\paragraph{Error criteria} We evaluate models using the mean squared error (MSE) between the benchmark solutions and the predictions of different models.

\begin{remark}
           The iPINNER employs an ensemble strategy, which can be computationally expensive. This overhead, however, can be alleviated through parallelization, as each ensemble member is trained independently. Furthermore, the incorporation of the evolutionary multi-objective optimization algorithm (NSGA-III) accelerates the convergence of each ensemble member relative to a standard PINN. As a result, while the computational cost of iPINNER is higher, it is not prohibitive. Table~\ref{tab:walltime_comparison} in revised manuscript (also attached below) list the wall time for training PINN and the proposed iPINNER. The wall time is recorded until the model reaches a convergence plateau.
        In addition, we performed a sensitivity analysis to compare the computational cost with respect to the ensemble size. The results, summarized in Table \ref{tab:sensitivity_analysis}, show that the optimal ensemble size is approximately $N=8$, which provides a balance between accuracy and computational efficiency.
        All experiments were performed on a linux machine with an NVIDIA H100 GPU (80G). 
        \begin{table}[htbp]
\centering
\caption{Wall time of training PINN and iPINNER.
}
\label{tab:walltime_comparison}
\begin{tabular}{lcc}
\toprule
\textbf{Method} & \textbf{Wall Time (s)} & \textbf{MSE} \\
\midrule
Standard PINN    & 205  & 0.0036 \\
\textbf{iPINNER} & \textbf{385 }& \textbf{0.0014} \\
\bottomrule
\end{tabular}
\end{table}
\end{remark}
    \begin{table}[H]
\centering
\caption{Sensitivity analysis for the ensemble size ($N$). The table shows the trade-off between the final Mean Squared Error (MSE) and the total wall-clock time. The chosen value, $N=8$, offers the best balance.}
\label{tab:sensitivity_analysis}
\begin{tabular}{ccc}
\toprule
\textbf{Ensemble Size ($N$)} & \textbf{Final MSE} & \textbf{Wall-Clock Time (s)} \\
\midrule
4                         & 0.0019              & 298                             \\
6 & 0.0016 &336\\
8                         & 0.0014              & 385                             \\
10                        & 0.0014             & 415                            \\
\bottomrule
\end{tabular}
\end{table}
}
 

\subsubsection{Forward Problem}
In the forward problem, the model is assumed to be not perfect which yields the model error. In particular, we assume that the viscosity term $v$ is different from its exact value which represents the model errors, i.e., we choose $\nu = 0.02/\pi$ which is different from its true value $\nu=0.01/\pi$. 
In this setting, we evaluate the performance of three models: ADAM-PINN, NSGA-PINN, and MoPONNEnKF.\\
\begin{table}[h!]
\centering
\begin{tabular}{|c|c|c|}
\hline
\textbf{Model} & \textbf{Noise Level} & \textbf{Testing Error} \\
\hline
\multirow{4}{*}{PINN} 
              & 0\% (no data) & 0.0036 \\
              & 20\% noise    & 0.0014 \\
              & 50\% noise    & 0.0029 \\
              & 80\% noise    & 0.0036 \\
\hline
\end{tabular}
\caption{Testing error of ADAM-PINN with different Gaussian noises in the presence of imperfect model.}
\label{tab:pinn_noise}
\end{table}
{Table~\ref{tab:pinn_noise} shows the testing errors of  ADAM-PINNs with different Gaussian noise in available observational data, with imperfect model. Because the imperfect model introduces the incorrect viscosity in Burgers equation, the PINN yields model errors. By combining noisy observation data with the model, iPINNER finds the optimal between the two. It is also noted that when the noise level is high, i.e., 80\%, both the model error and observational error dominate and hence the iPINNER's accuracy diminishes.
}

\begin{table}[H]
    \centering
    \begin{tabular}{lcccc}
        \toprule
        Model & 0\% (no data) & 20\% data noise & 50\% data noise & 80\% data noise \\
        \midrule
        ADAM-PINN &0.0036& 0.0014 & 0.0029 & 0.0036 \\
        NSGA-III-PINN &0.0030& 0.0016 & 0.0024 & 0.0033 \\
        \textbf{iPINNER} &\textbf{N/A}& \textbf{0.0006} & \textbf{0.0009} & \textbf{0.0014} \\
        \bottomrule
    \end{tabular}
    \caption{Mean square errors (MSEs) of three different models, ADAM-PINN, NSGA-III-PINN, iPINNER, with different noise levels in the one-dimensional Burgers Equation forward test problem.}
    \label{tab:Burgers Forward testing error}
\end{table}
Table \ref{tab:Burgers Forward testing error} shows the mean square error (MSE) in the forward problem setting for (1). ADAM-PINN, (2). NSGA-III-PINN, and (3). iPINNER with different noise level of obervational data.
{
It is shown that, when the observational noise level is $20\%$ and $50\%$, iPINNER are at least one order more accurate than ADAM-PINN and NSGA-III-PINN. Even when the observational data become substantially noisier ($80\%$ noise level), it remains at least twice as accurate as the other two methods.}
This is not surprising because both ADAM-PINN and NSGA-III-PINN use noise-contaminated observations directly in their loss functions during training, without an explicit noise-filtering mechanism. In contrast, the iPINNER framework incorporates an additional "purification" step through the Kalman filter, thereby effectively mitigating the impact of observational noise.
Nonetheless, the NSGA-III-PINN exhibits slightly better robustness than ADAM-PINN, as the NSGA-III algorithm more effectively balances multiple objectives within the loss function, thereby partially mitigating the adverse effects of noisy observations. 
\begin{table}[H]
    \centering
    \begin{tabular}{lcccc}
        \toprule
        Model & Residual loss & Boundary loss & Observation loss & Testing error \\
        \midrule
        ADAM-PINN & 0.0005 & 0.0002 & 0.0381 & 0.0036\\
        NSGA-III-PINN &0.0002  &0.0002  &0.0320  &0.0033\\
        \textbf{iPINNER} & \textbf{0.0001} & $\boldsymbol{7.25 \times 10^{-5}}$
 & \textbf{0.0020} &  \textbf{0.0014}
  \\
        \bottomrule
    \end{tabular}
    \caption{Mean square errors (MSEs) of three different models, ADAM-PINN, NSGA-III-PINN, iPINNER, with $80\%$ noise levels in the one-dimensional Burgers Equation forward test problem.}
    \label{tab:Burgers Forward}
\end{table}
{Table~\ref{tab:Burgers Forward} shows that the NSGA-III algorithm effectively balances the loss terms in the PINN to comparable orders of magnitude within a reasonable training time. As a result, during training, the proposed iPINNER method achieves relatively low residuals for the PDE residual loss, boundary loss, and data loss—performance that neither ADAM-PINN nor NSGA-III-PINN alone can attain.
}

\begin{figure}[H]
    \centering
    \includegraphics[width=.32\linewidth]{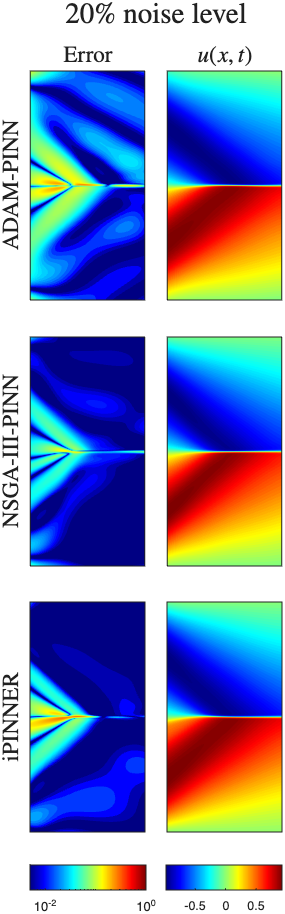}
    \includegraphics[width=.32\linewidth]{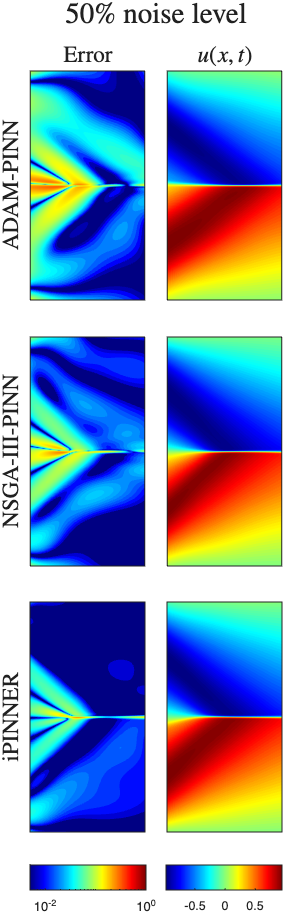}
    \includegraphics[width=.32\linewidth]{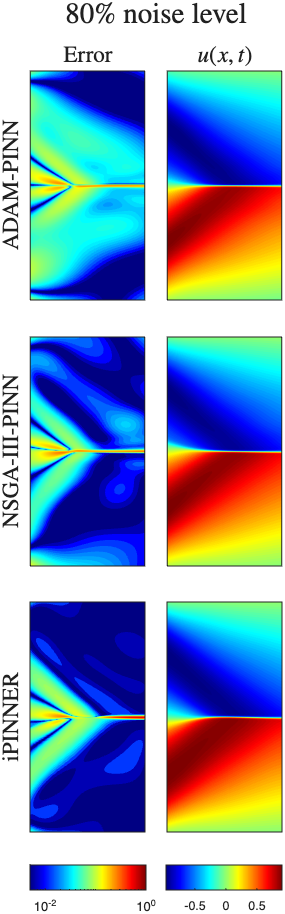}
    \caption{Forward problem solutions (top) and errors (bottom) for comparison of ADAM-PINN, NSGA-III-PINN and iPINNER solution for one dimensional Burgers equation; observation data noise level are chosen to be $20\%$, $50\%$, and $80\%$ respectively.}
    \label{fig:nsga-kf-20}
\end{figure}

Figure~\ref{fig:nsga-kf-20} shows the spatiotemporal solutions and corresponding errors obtained using (1) ADAM-PINN, (2) NSGA-III-PINN, and (3) iPINNER, under different observational noise levels. Consistent with the findings summarized in Table~\ref{tab:Burgers Forward testing error}, iPINNER is the most accurate model among the three approaches, with its accuracy becoming more remarkable as the noise level increases.

\subsubsection{Inverse Problem}
In the inverse problem, we assume that the diffusion term is unknown, i.e., the diffusion coefficient $\nu$ in equation \eqref{eq:burgers}, is not known, and its initial estimate are chosen randomly in a certain range from $0$ to $1$. 
The diffusion coefficient is considered to be a trained parameter in the PINN framework  \textcolor{blue}{and} we use data with varying noise levels—similar to those in the forward problem—to train each PINN model. 
The other neural network settings are the same as in Section \ref{sec:nn-setting}. 

\begin{figure}[H]
    \centering
    \includegraphics[width=.32\linewidth]{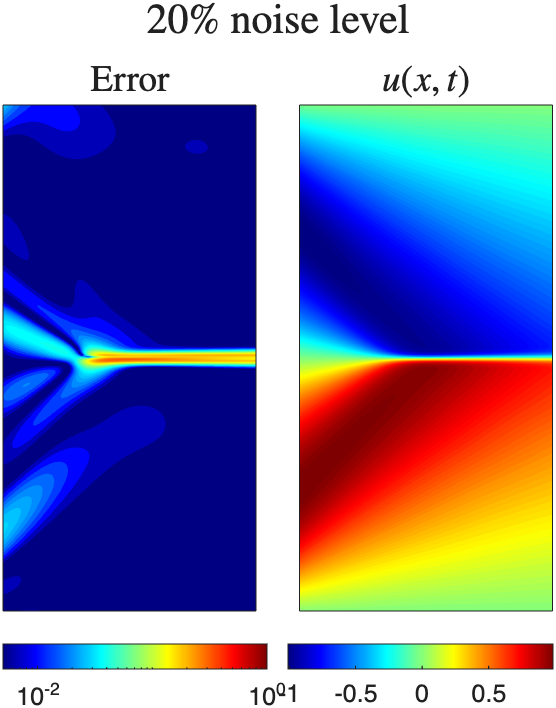}
    \includegraphics[width=.32\linewidth]{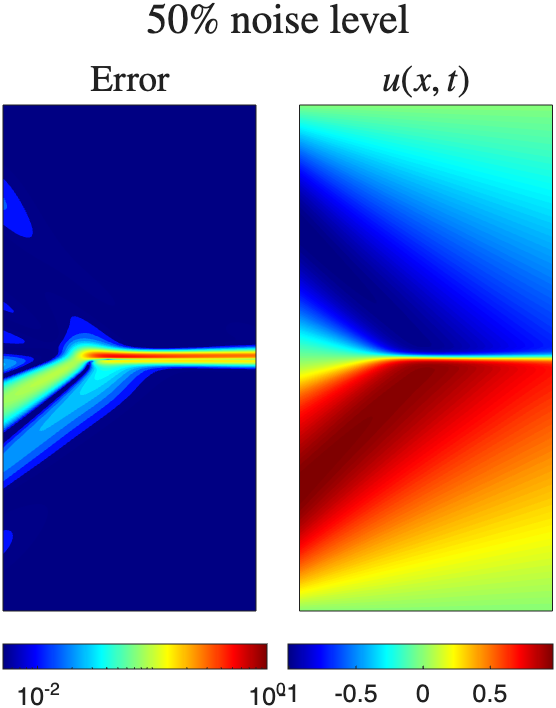}
    \includegraphics[width=.32\linewidth]{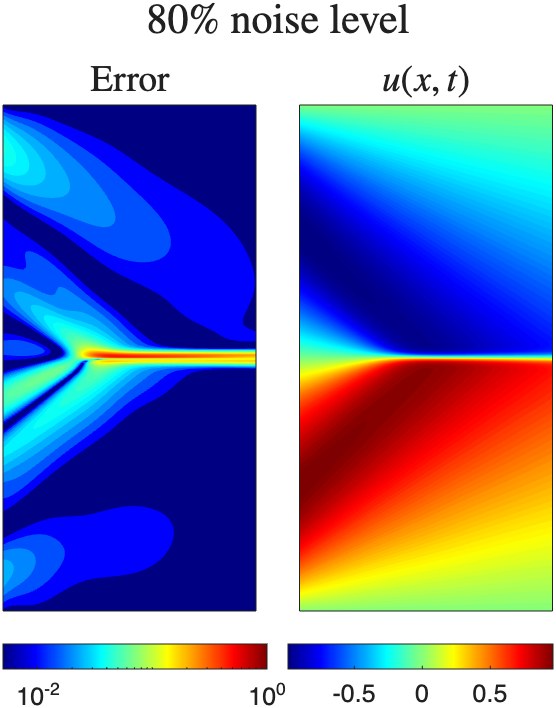}
    \caption{Inverse problem errors (left) and solutions (right) for iPINNER solution for one dimensional Burgers equation; observation data noise level are chosen to be $20\%$, $50\%$, and $80\%$ respectively.}
    \label{fig:nsga-kf-inverse-20}
\end{figure}

Table \ref{tab:Burgers Inverse testing error}
shows the estimated diffusion parameters with $L^1$ errors for different models. 
It is shown that for all different observational uncertainties, the proposed iPINNER gives the best estimate. Specifically when the observational uncertainty is small and medium, i.e., $\eta=20\%$ and $50\%$ respectively, the proposed iPINNER gives the much better estimate when comparing to PINNs with ADAM optimizer and with NSGA-III optimizer; and between the two, the later is slightly better. Yet, when the observational uncertainty is large, i.e., $\eta = 80\%$, the proposed iPINNER gives similar inaccurate results as the other two. This is because when the data is noisy enough, the diffusion coefficient which is the trainable parameter together with PINN's weights converges to the wrong that still can represent the limited noisy data. 
Figure \ref{fig:nsga-kf-inverse-20} shows  errors (left) and solutions (right) for iPINNER solution for one dimensional Burgers equation with different observation data noise level are chosen to be $20\%$, $50\%$, and $80\%$. It is consistent with the results shown in Table \ref{tab:Burgers Inverse testing error}.
\begin{table}[H]
    \centering
    \renewcommand{\arraystretch}{1.2} %
    \begin{tabular}{ccccc}
        \toprule
        Noise level &Model & $\nu$ estimation &  $L^1$ error  \\

        \midrule
        \multirow{3}{*}{20\%} &  ADAM-PINN    & 0.0095  &  0.00632  \\
                              &  NSGA-III-PINN    & 0.00745  &  0.00427  \\
                              &   \textbf{iPINNER} & \textbf{0.00546}  &  \textbf{0.00287} \\
        \midrule
        \multirow{3}{*}{50\%} & ADAM-PINN     &0.0118   &  0.00862  \\
                              & NSGA-III-PINN   &0.00788   &  0.0047  \\
                              &  \textbf{iPINNER} & \textbf{0.00648} & \textbf{0.0033}  \\
        \midrule                      
        \multirow{3}{*}{80\%} & ADAM-PINN    &0.0094  &   0.00622   \\
                              & NSGA-III-PINN     & 0.00902  &  0.00584   \\
                              & \textbf{iPINNER} & \textbf{0.00844}  & \textbf{0.00526}\\
        \bottomrule
    \end{tabular}
    \caption{Comparison of ADAM-PINN, NSGA-III-PINN and the proposed iPINNER for estimating the viscosity term $\nu$ in the Burgers equation. For reference, the true value of the viscosity term is $\nu=0.01/\pi\approx 0.00318$.}
    \label{tab:Burgers Inverse testing error}
\end{table}

\subsection{Time-fractional Mixed Diffusion-Wave Equations (TFMDWEs) \label{sec:frac}}
In this section, we consider the \textit{time-fractional mixed diffusion-wave equations} (TFMDWEs), which generalize classical diffusion and wave equations by incorporating fractional-order time derivatives \cite{liu2019alternating,du2021temporal,lu2025fpinn}. The TFMDWEs with Dirichlet boundary conditions are defined as follows:
\begin{align}
& D_t^{\alpha} u(x,t) = \frac{\partial^2 u}{\partial x^2} + f(x,t), \quad t \in [0,1],\; x \in \Omega \equiv [0,\pi],\\[6pt]
&u(x,t) = 0,\quad \forall x \in \partial\Omega,\\[6pt]
& u(x, 0) = 0, \quad x \in \Omega,
\end{align}
where the fractional order \(\alpha\) yields:
\begin{align}
\alpha \in [0,1], \quad \forall t \in [0,1],
\end{align}
and the forcing term \(f(x,t)\) is explicitly defined as:
\begin{align}
f(x,t) = \frac{\Gamma(4)}{\Gamma(4-\alpha)}\, t^{3-\alpha}\sin(x) + t^3\sin(x),
\end{align}
with $\Gamma(\cdot)$ denoting the Gamma function. This equation introduces fractional-order temporal dynamics, combining features of both diffusion and wave phenomena. The fractional order $\alpha$ controls the transition between diffusive and wave-like behaviors, making TFMDWEs particularly useful in modeling complex systems exhibiting anomalous transport and non-local temporal interactions.
\subsubsection{PINN's Settings \label{sec:nn-setting}}

\begin{figure}[H]
    \centering
    \resizebox{\textwidth}{!}{  
    \begin{minipage}{\textwidth}
        \centering    \begin{subfigure}[b]{0.305\linewidth}
    \centering
        \includegraphics[width=\linewidth]{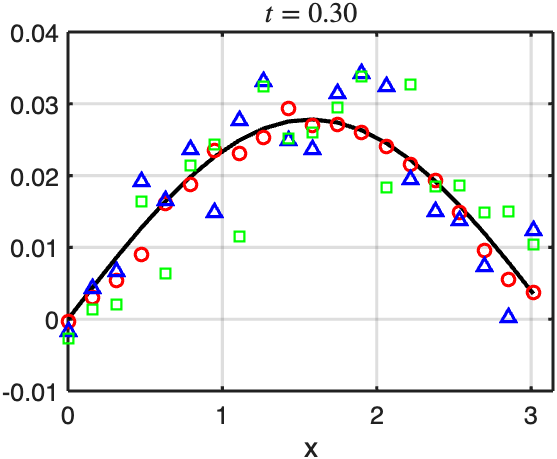}
        \caption{}
    \end{subfigure}
    \begin{subfigure}[b]{0.3\linewidth}
    \centering
        \includegraphics[width=\linewidth]{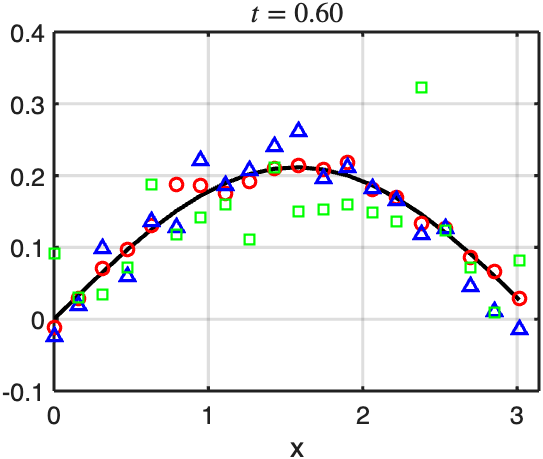}
        \caption{}
    \end{subfigure}
    \begin{subfigure}[b]{0.3\linewidth}
    \centering
        \includegraphics[width=\linewidth]{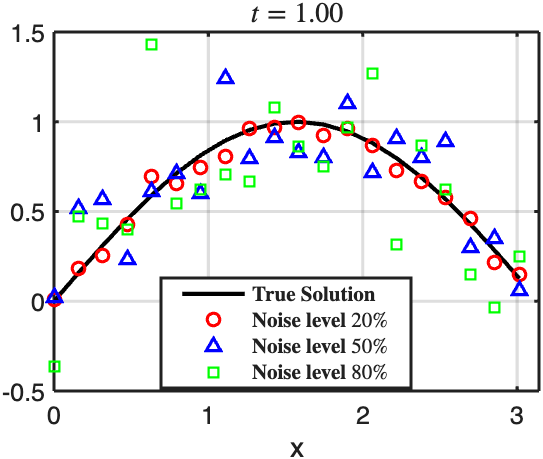}
        \caption{}
    \end{subfigure}
        \end{minipage}
    }
    \caption{
Observation data with different noise levels at time instances (a)~$t = 0.3$, (b)~$t = 0.6$, and (c)~$t = 1$ for the time-fractional mixed diffusion-wave equations (TFMDWEs).
    }
    \label{fig:fpde-noise}
\end{figure}

\paragraph{Observational data}
To generate observational data used in PINN framework, we choose $100$ data points evenly for the initial and boundary conditions and $10^4$ interior points. Figure~\ref{fig:fpde-noise} shows the different noise level observation at different time instance for the time-fractional
mixed diffusion-wave equations (TFMDWEs).
\paragraph{Neural network architecture} 
In order to make a fair comparison, with different optimizers, the PINN employs a consistent deep neural network architecture comprising 2 hidden layers with 50 neurons in each layer.
During the training phase, models corresponding to each optimizer were collected until the training loss converged below a prescribed threshold $\epsilon$.
Specifically, training ADAM-PINN requires $5000$ epochs to achieve convergence, whereas the NSGA-III-PINN reaches convergence within $4$ generations. The proposed iPINNER requires of $3$ generations with $2000$ epochs to ensure the convergence of training loss.

\subsubsection{Forward Problem}
\begin{figure}[H]
    \centering
    \resizebox{0.7\textwidth}{!}{  
    \begin{minipage}{\textwidth}
        \centering    
        \begin{subfigure}[b]{0.48\linewidth}
    \centering
        \includegraphics[width=\linewidth]{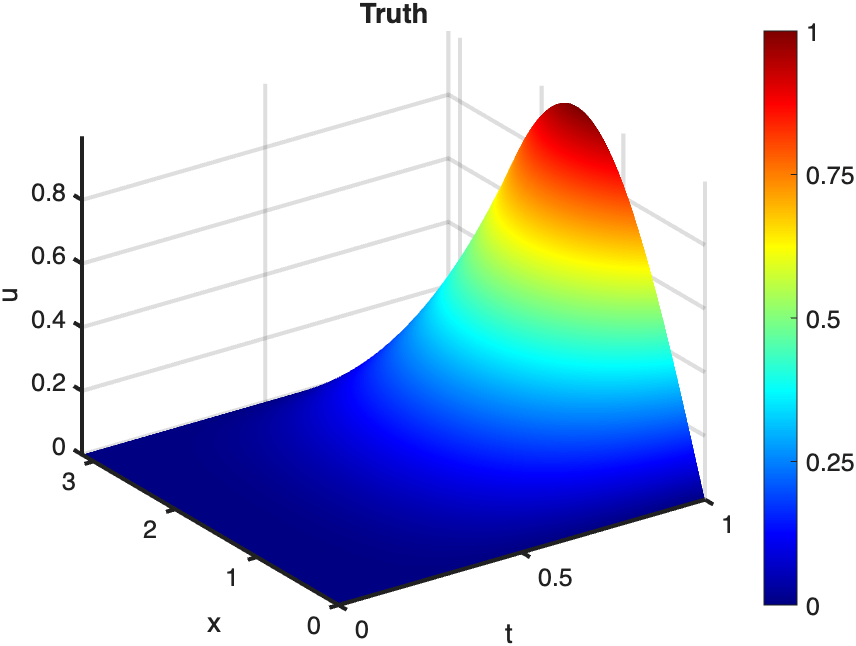}
        \caption{}
    \end{subfigure}
    \hfill
    \begin{subfigure}[b]{0.48\linewidth}
    \centering
        \includegraphics[width=\linewidth]{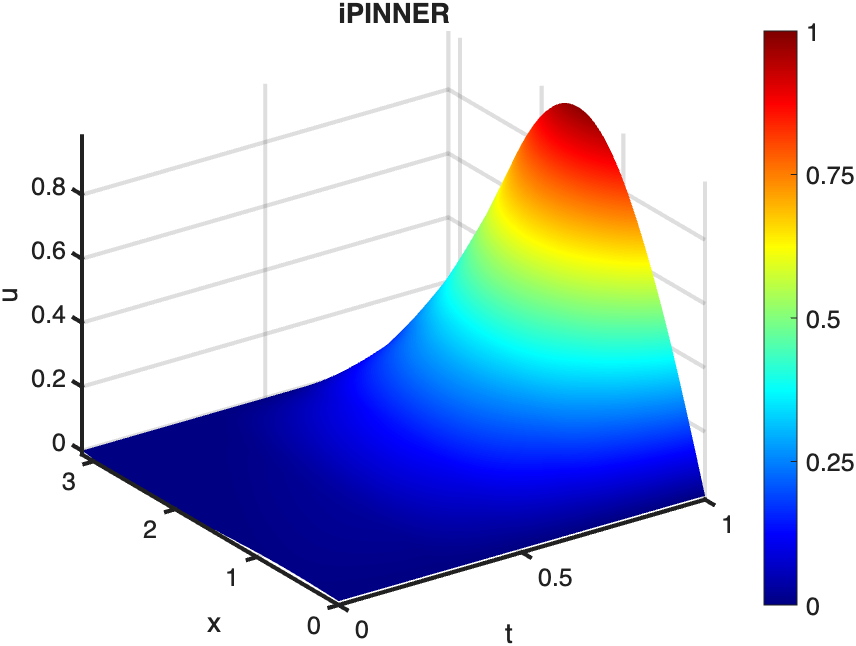}
        \caption{}
    \end{subfigure}
    \begin{subfigure}[b]{0.48\linewidth}
    \centering
        \includegraphics[width=\linewidth]{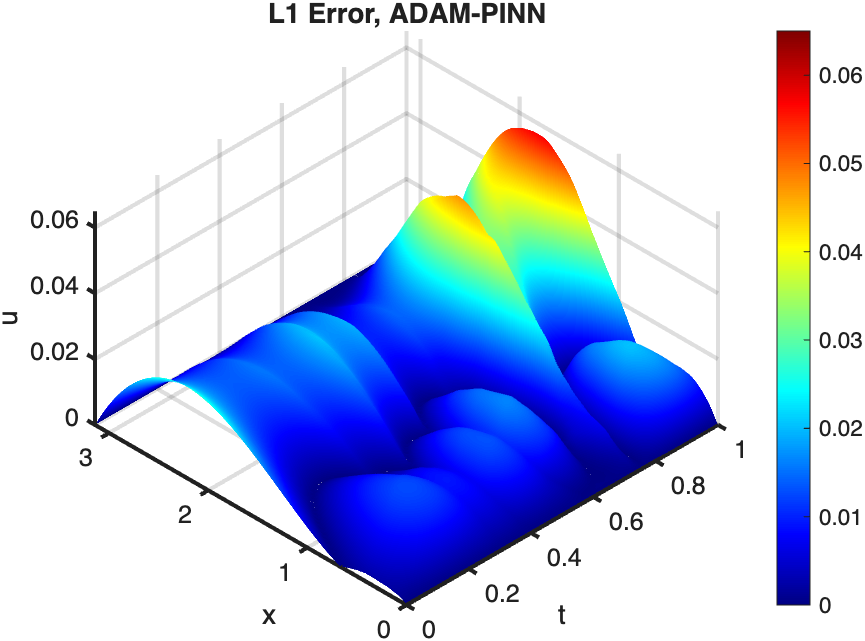}
        \caption{}
    \end{subfigure}
    \hfill
    \begin{subfigure}[b]{0.48\linewidth}
    \centering
        \includegraphics[width=\linewidth]{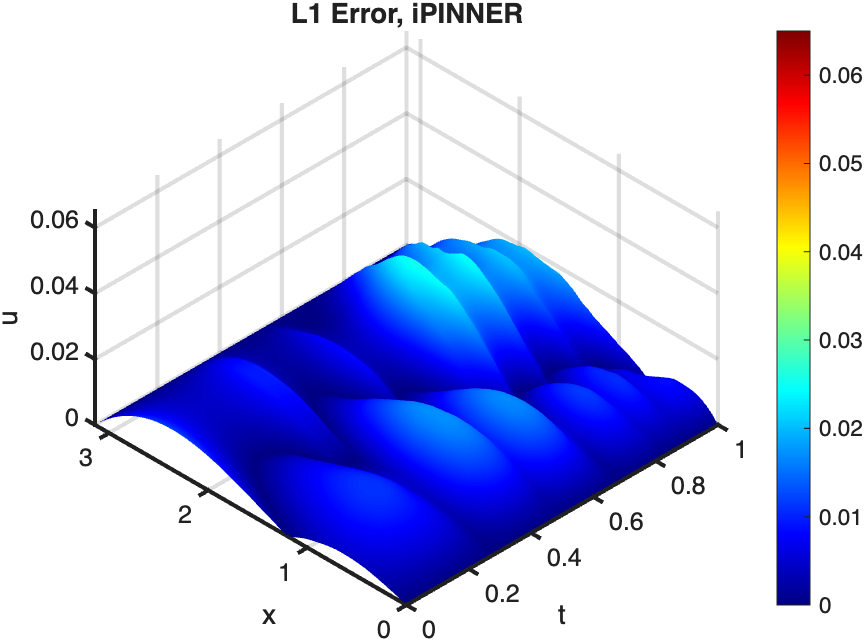}
        \caption{}
    \end{subfigure}
        \end{minipage}
    }
    \caption{
        Comparison of (a) the true solution and (b) the iPINNER solution as well as the $L_1$ errors of (c) the ADAM-PINN and (d) the iPINNER for  the time-fractional mixed diffusion-wave equations (TFMDWEs) in forward problem. Observation noise level is $20\%$.
    }
    \label{fig:fpde-forward-20}
\end{figure}

\begin{figure}[H]
    \centering
    \resizebox{0.7\textwidth}{!}{  
    \begin{minipage}{\textwidth}
        \centering    
    \begin{subfigure}[b]{0.48\linewidth}
    \centering
        \includegraphics[width=\linewidth]{TFMDWEs_true_forward_noise20.png}
        \caption{}
    \end{subfigure}
    \hfill
    \begin{subfigure}[b]{0.48\linewidth}
    \centering
        \includegraphics[width=\linewidth]{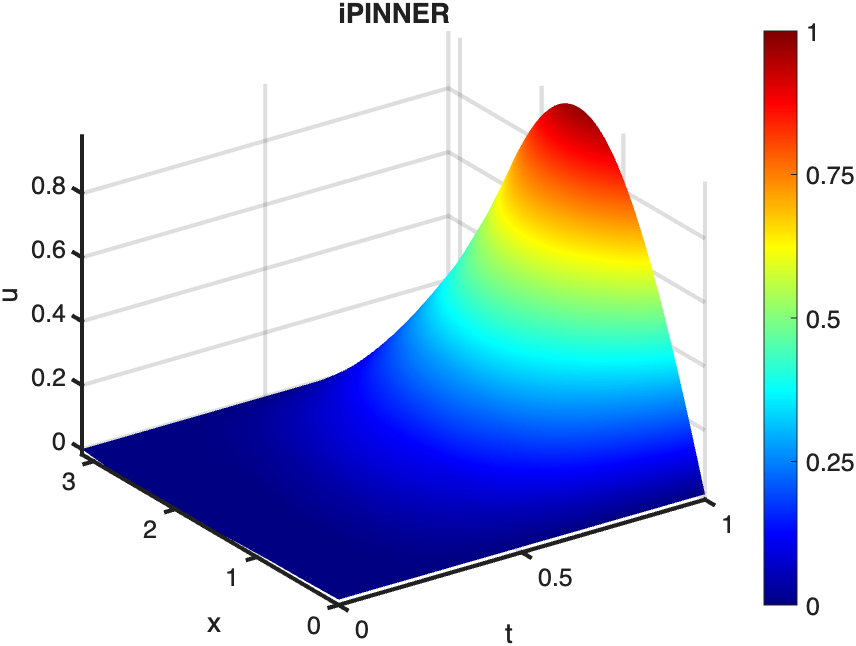}
        \caption{}
    \end{subfigure}
    \begin{subfigure}[b]{0.48\linewidth}
    \centering
        \includegraphics[width=\linewidth]{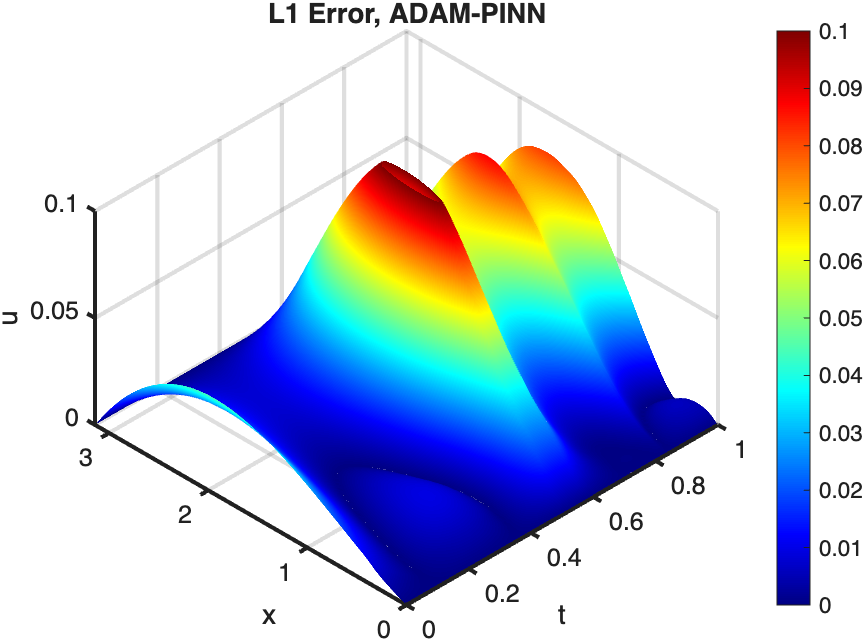}
        \caption{}
    \end{subfigure}
    \hfill
    \begin{subfigure}[b]{0.48\linewidth}
    \centering
        \includegraphics[width=\linewidth]{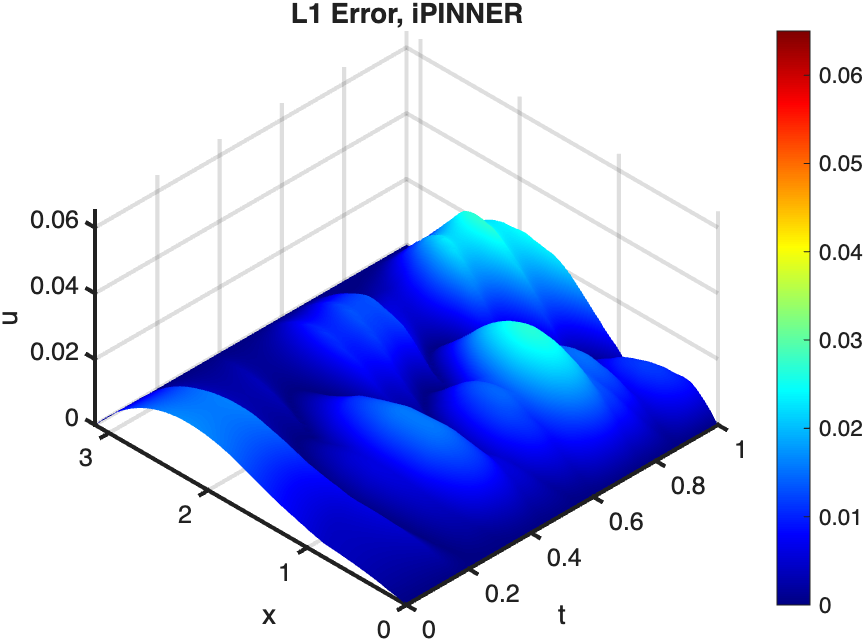}
        \caption{}
    \end{subfigure}
        \end{minipage}
    }
    \caption{
      Comparison of (a) the true solution and (b) the iPINNER solution as well as the $L_1$ errors of (c) the ADAM-PINN and (d) the iPINNER for  the time-fractional mixed diffusion-wave equations (TFMDWEs) in forward problem.  Observation noise level is $50\%$.
    }
    \label{fig:fpde-forward-50}
\end{figure}

\begin{figure}[H]
    \centering
    \resizebox{0.7\textwidth}{!}{  
    \begin{minipage}{\textwidth}
        \centering
        \begin{subfigure}[b]{0.48\linewidth}
            \centering
            \includegraphics[width=\linewidth]{TFMDWEs_true_forward_noise20.png}
            \caption{}
        \end{subfigure}
        \hfill
        \begin{subfigure}[b]{0.48\linewidth}
            \centering
            \includegraphics[width=\linewidth]{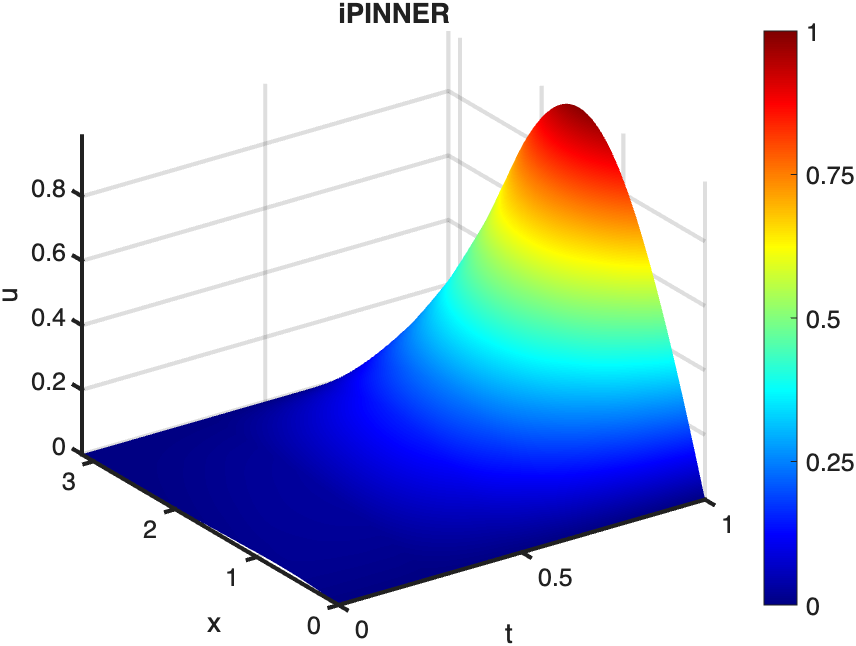}
            \caption{}
        \end{subfigure}

        \vspace{2mm}

        \begin{subfigure}[b]{0.48\linewidth}
            \centering
            \includegraphics[width=\linewidth]{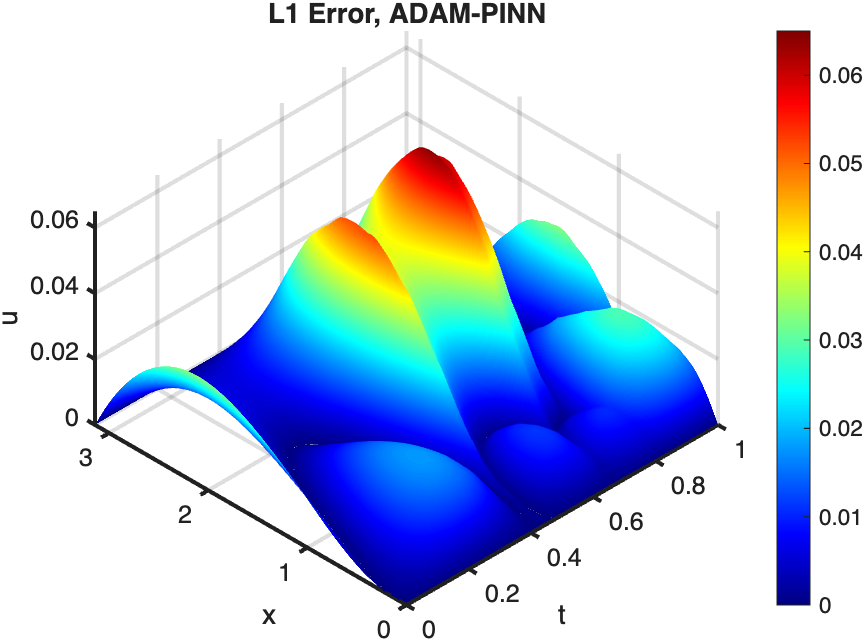}
            \caption{}
        \end{subfigure}
        \hfill
        \begin{subfigure}[b]{0.48\linewidth}
            \centering
            \includegraphics[width=\linewidth]{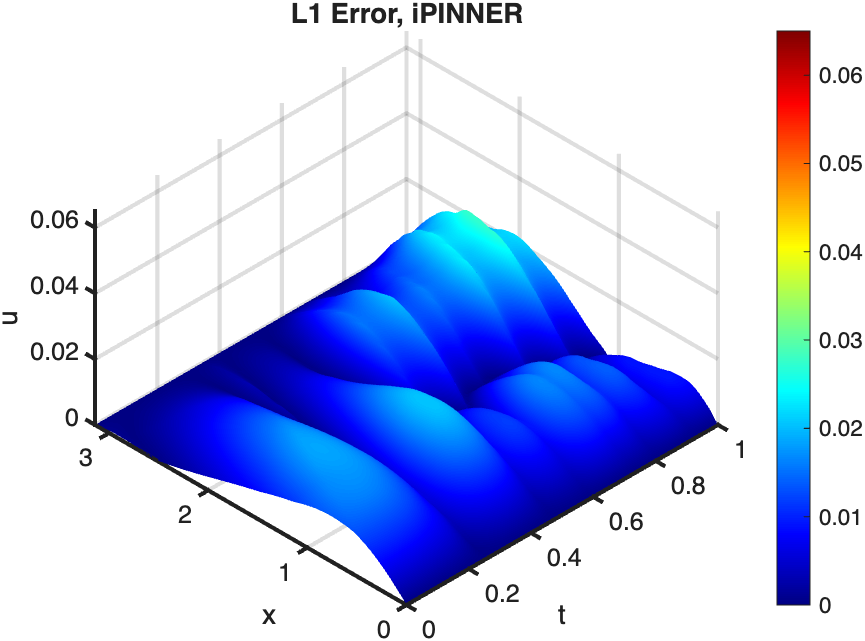}
            \caption{}
        \end{subfigure}
    \end{minipage}
    }

    \caption{
          Comparison of (a) the true solution and (b) the iPINNER solution as well as the $L_1$ errors of (c) the ADAM-PINN and (d) the iPINNER for  the time-fractional mixed diffusion-wave equations (TFMDWEs) in forward problem.  Observation noise level is $80\%$.
    }
    \label{fig:fpde-forward-80}
\end{figure}

\begin{figure}
    \centering
      \begin{subfigure}[b]{0.3\linewidth}
            \centering
    \includegraphics[width=\linewidth]{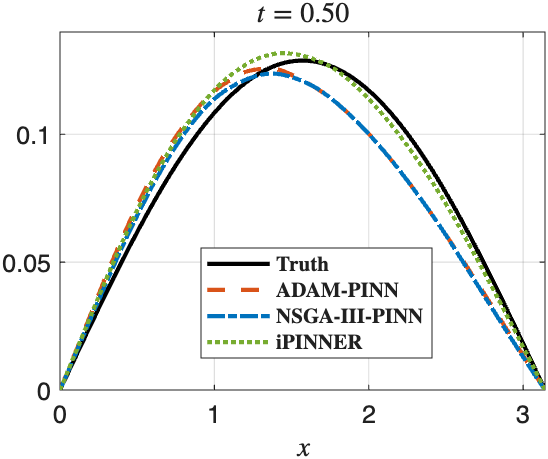}
    \includegraphics[width=.96\linewidth]{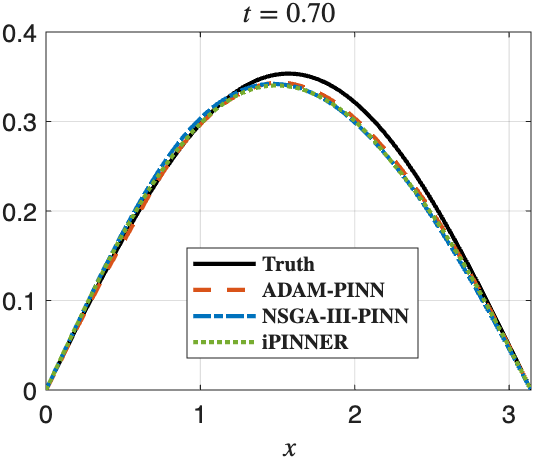}
                \caption{$20\%$ noise level }
        \end{subfigure}
              \begin{subfigure}[b]{0.3\linewidth}
            \centering
    \includegraphics[width=\linewidth]{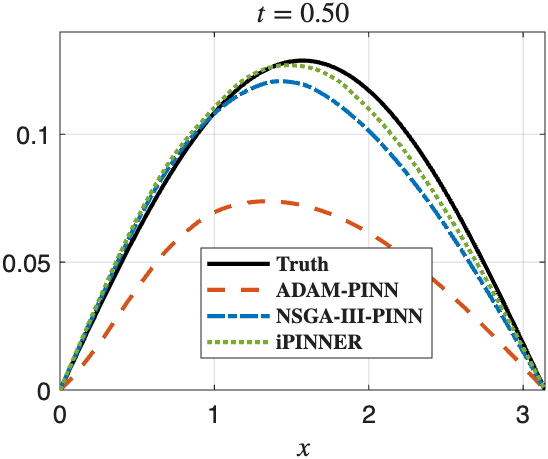}
    \includegraphics[width=.96\linewidth]{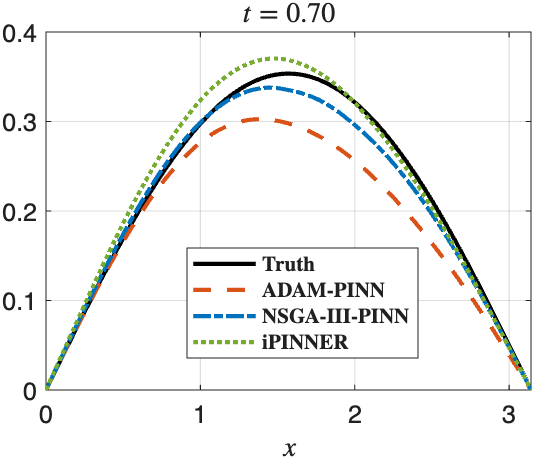}
                \caption{$50\%$ noise level }
        \end{subfigure}
              \begin{subfigure}[b]{0.3\linewidth}
            \centering
    \includegraphics[width=\linewidth]{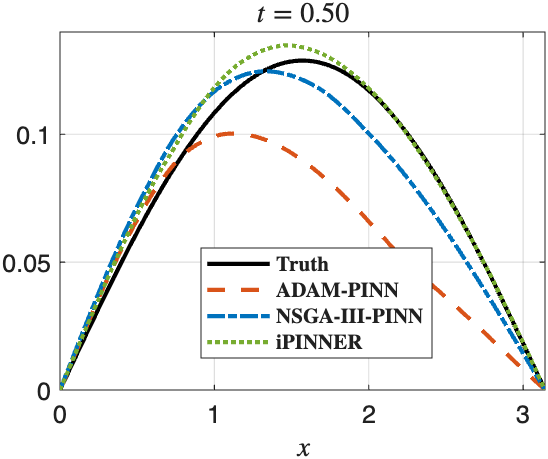}
    \includegraphics[width=.96\linewidth]{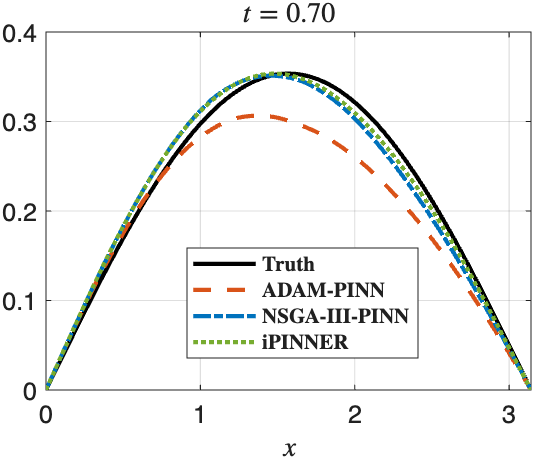}
                \caption{$80\%$ noise level }
        \end{subfigure}
    \caption{Comparison at different time instances (top: $t=0.5$, bottom: $t=0.7$) between the true solution (black) and forward problem solutions obtained using ADAM-PINN (red), NSGA-III-PINN (blue), and iPINNER (green) for the time-fractional mixed diffusion-wave equations (TFMDWEs) under varying noise levels: Panel (a) — $20\%$ noise, Panel (b) — $50\%$ noise, and Panel (c) — $80\%$ noise. }
    \label{fig:fpde-curve-forward}
\end{figure}

{Similar to the Burgers equation test case, we assume that only an imperfect model is available. In particular, for TFMDWEs, the source term $f$  is assumed to be inaccurate by adding 50\% Gaussian noise and thereby mimicking the model errors.}
\begin{table}[H]
\centering
\begin{tabular}{|c|c|c|}
\hline
\textbf{Model} & \textbf{Noise Level} & \textbf{Testing Error} \\
\hline
\multirow{4}{*}{PINN} 
              & 0\% (no data) & 0.0999 \\
              & 20\% noise    & 0.0878 \\
              & 50\% noise    & 0.0895 \\
              & 80\% noise    & 0.1089 \\
\hline
\end{tabular}
\caption{Testing error of PINN (MAE) under different levels of Gaussian noise with incorrect viscosity.}
\label{tab:pinn_noise_FPDE}
\end{table}
{Table~\ref{tab:pinn_noise_FPDE} shows the testing error of PINN model with different Gaussian noises in observational data. With the imperfect model, the PINN leads to inaccurate prediction without data. Incorporating observational data with low-level noise (20\%) can improve the correction of model predictions. However, when the observational data are highly contaminated by noise, the predictive accuracy of the model deteriorates.
}
\begin{table}[H]
    \centering
    \begin{tabular}{lcccc}
        \toprule
        Model &0\% (no data)& 20\% data noise & 50\% data noise & 80\% data noise \\
        \midrule
        ADAM-PINN &0.0999 &0.0878  &0.0885  &0.1089  \\
        NSGA-III-PINN &0.0889 &0.0860  &0.0875  & 0.0887  \\
          \textbf{iPINNER} &\textbf{N/A}&\textbf{0.0850}  &\textbf{0.0865}  & \textbf{0.0872} \\
        \bottomrule
    \end{tabular}
    \caption{Mean absolute errors (MAEs) of three different models, ADAM-PINN, NSGA-III-PINN, iPINNER, with different noise levels in the time-fractional mixed diffusion-wave equations (TFMDWEs).}
    \label{tab:fpde-forward}
\end{table}
Table \ref{tab:fpde-forward} shows the mean absolute error (MAE) in the forward problem setting for (1). ADAM-PINN, (2). NSGA-III-PINN, and (3). iPINNER with different noise level of observational data.
It is shown that in all cases, iPINNER is at least more accurate than ADAM-PINN and NSGA-III-PINN. 
Nonetheless, the NSGA-III-PINN exhibits slightly better robustness than ADAM-PINN, as the NSGA-III algorithm more effectively balances multiple objectives within the loss function, thereby partially mitigating the adverse effects of noisy observations. This is also reflected in the figures \ref{fig:fpde-forward-20}--\ref{fig:fpde-forward-80}.
\begin{table}[H]
    \centering
    \begin{tabular}{lcccc}
        \toprule
        Model & Residual loss & Boundary loss & Observation loss & Testing error \\
        \midrule
        ADAM-PINN & 0.1847 & 0.0182 & 0.0490 & 0.0232\\
        NSGA-III-PINN &0.1742  &0.0155  &0.0475  &0.0219\\
        \textbf{iPINNER} & \textbf{0.1784} & \textbf{0.0166}
 & \textbf{0.0449} &  \textbf{0.0203}
  \\
        \bottomrule
    \end{tabular}
    \caption{Mean square errors (MSEs) of three different models, ADAM-PINN, NSGA-III-PINN, iPINNER, with $80\%$ noise levels in the one-dimensional Burgers Equation forward test problem.}
    \label{tab:FPDE Forward}
\end{table}
{Table~\ref{tab:FPDE Forward} compares the mean square error of three different models, ADAM-PINN, NSGA-III-PINN, and iPINNER, with 80\% Gaussian noise in the forward problem. The results show that iPINNER consistently obtains the lowest errors across all loss components, i.e., PDE residual, boundary, and observation losses. }
Figure \ref{fig:fpde-curve-forward} shows at different time instances (top: $t = 0.5$, bottom: $t = 1$), the true solution (black) and forward problem
solutions obtained using ADAM-PINN (red), NSGA-III-PINN (blue), and iPINNER (green) for the time-fractional mixed diffusion-wave equations (TFMDWEs) with different noise levels. The results show that iPINNER (green) outperforms both ADAM-PINN (red) and NSGA-III-PINN (blue) in terms of accuracy. Moreover, its advantage becomes even more pronounced at higher noise levels.




{
In Figure \ref{fig:fpde-fem}, we show the finite element method (FEM) results with different mesh sizes, which demonstrate good agreement for the forward solution compared with our iPINNER. However, FEM is primarily restricted to solving forward problems. While FEM itself does not require data, with available observational data, it requires additional data assimilation techniques. On the other hand iPINNER can simultaneously addresses both forward and inverse problem settings with available noisy data. 

\begin{figure}[H]
    \centering
    \resizebox{\textwidth}{!}{  
    \begin{minipage}{\textwidth}
        \centering    
    \begin{subfigure}[b]{0.32\linewidth}
    \centering
        \includegraphics[width=\linewidth]{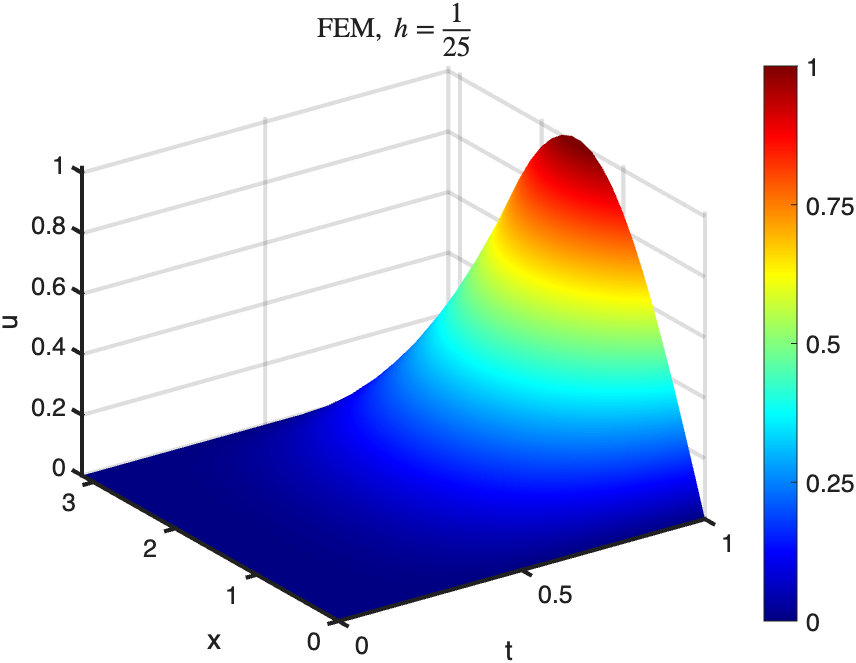}
        \caption{}
    \end{subfigure}
    \hfill
    \begin{subfigure}[b]{0.32\linewidth}
    \centering
        \includegraphics[width=\linewidth]{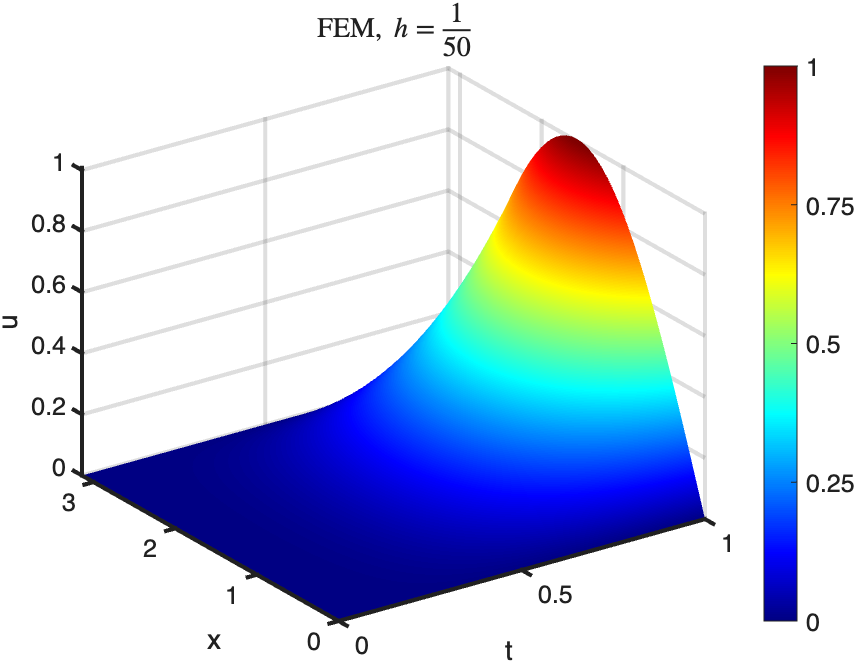}
        \caption{}
    \end{subfigure}
    \begin{subfigure}[b]{0.32\linewidth}
    \centering
        \includegraphics[width=\linewidth]{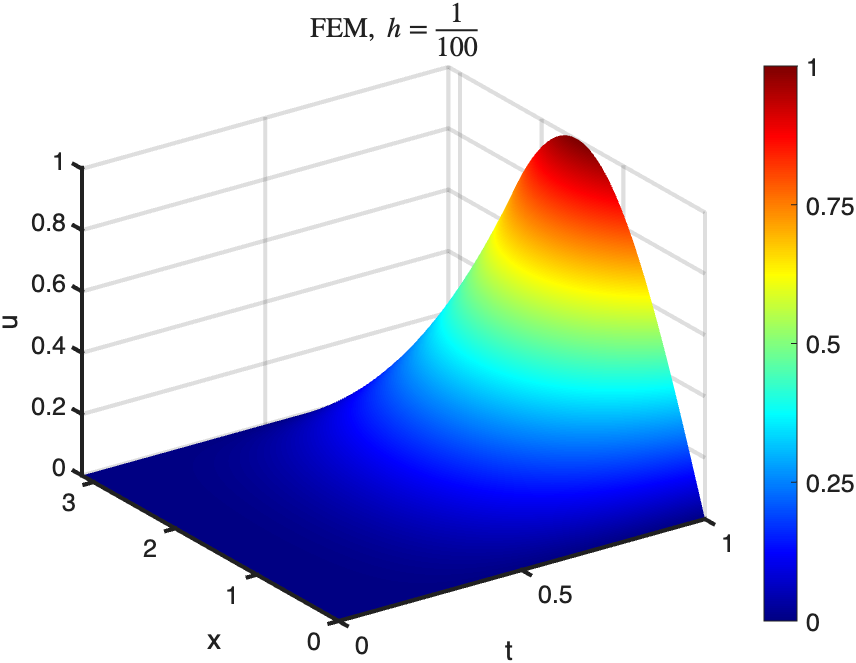}
        \caption{}
    \end{subfigure}
    \hfill
    \begin{subfigure}[b]{0.32\linewidth}
    \centering
        \includegraphics[width=\linewidth]{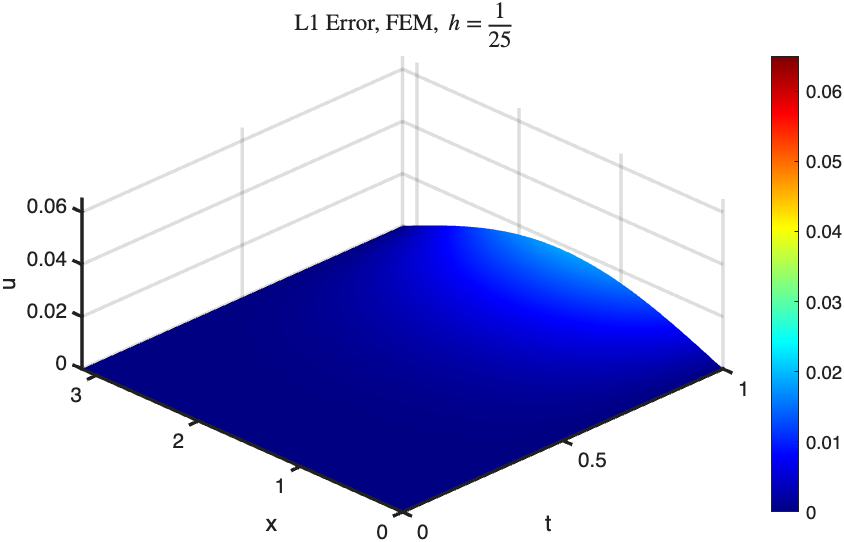}
        \caption{}
    \end{subfigure}
        \hfill
    \begin{subfigure}[b]{0.32\linewidth}
    \centering
        \includegraphics[width=\linewidth]{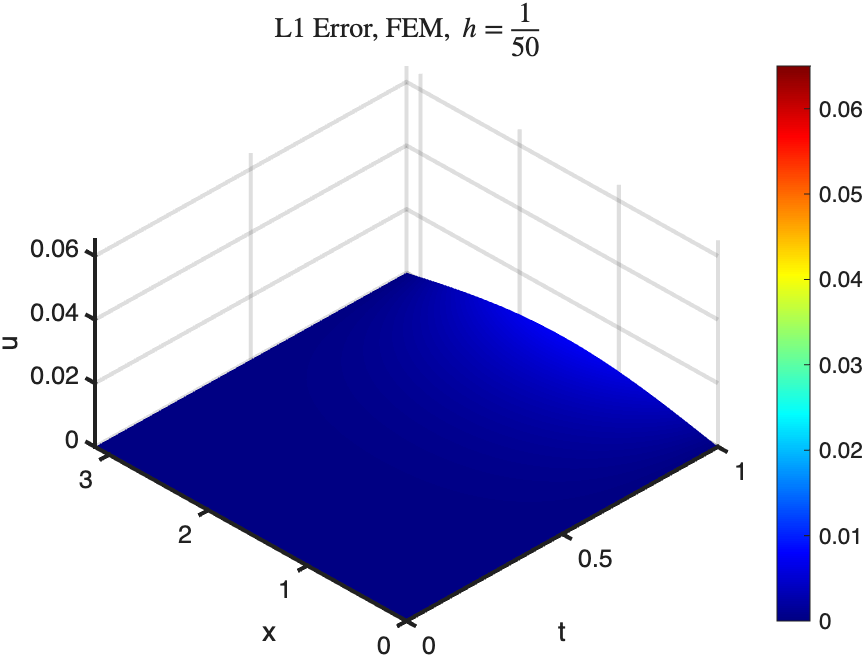}
        \caption{}
    \end{subfigure}
        \hfill
    \begin{subfigure}[b]{0.32\linewidth}
    \centering
        \includegraphics[width=\linewidth]{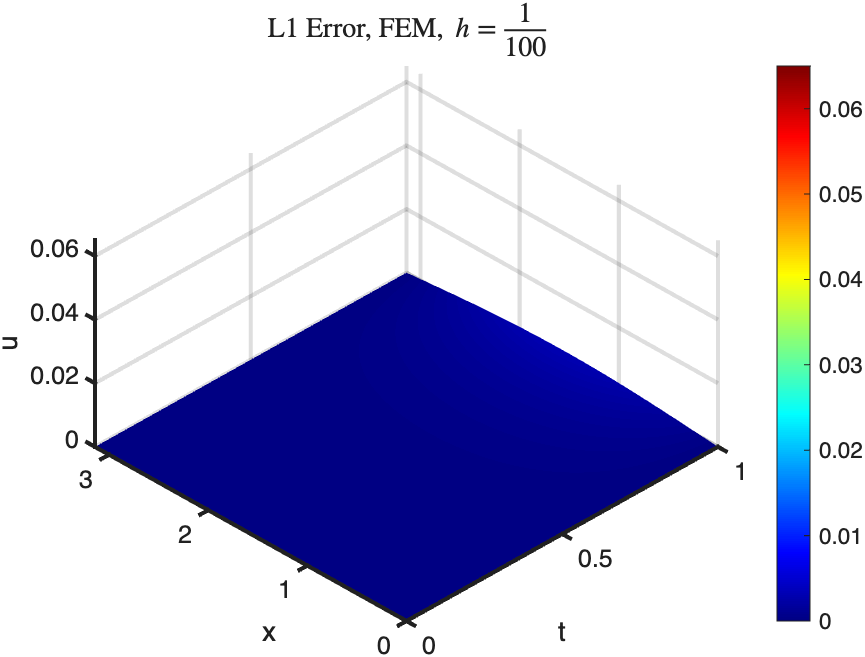}
        \caption{}
    \end{subfigure}
        \end{minipage}
    }
    \caption{
      Comparison of finite element method (FEM) solution with different mesh size for  the time-fractional mixed diffusion-wave equations (TFMDWEs) in forward problem.  
    }
    \label{fig:fpde-fem}
\end{figure}

}

\subsubsection{Inverse Problem}
\begin{figure}[H]
    \centering
    \resizebox{0.7\textwidth}{!}{  
    \begin{minipage}{\textwidth}
        \centering    \begin{subfigure}[b]{0.48\linewidth}
    \centering
        \includegraphics[width=\linewidth]{TFMDWEs_true_forward_noise20.png}
        \caption{}
    \end{subfigure}
    \hfill
    \begin{subfigure}[b]{0.48\linewidth}
    \centering
        \includegraphics[width=\linewidth]{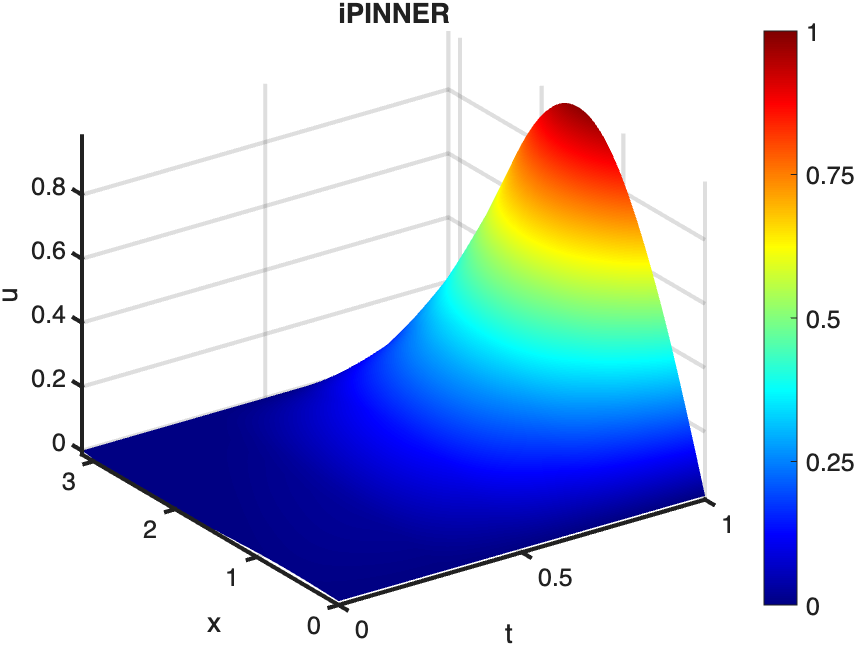}
        \caption{}
    \end{subfigure}
    \begin{subfigure}[b]{0.48\linewidth}
    \centering
        \includegraphics[width=\linewidth]{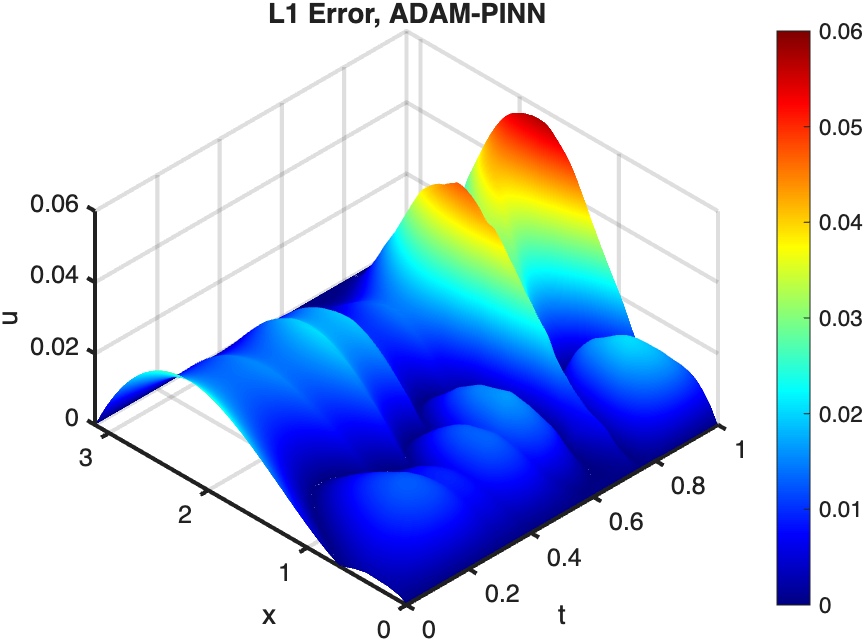}
        \caption{}
    \end{subfigure}
    \hfill
    \begin{subfigure}[b]{0.48\linewidth}
    \centering
        \includegraphics[width=\linewidth]{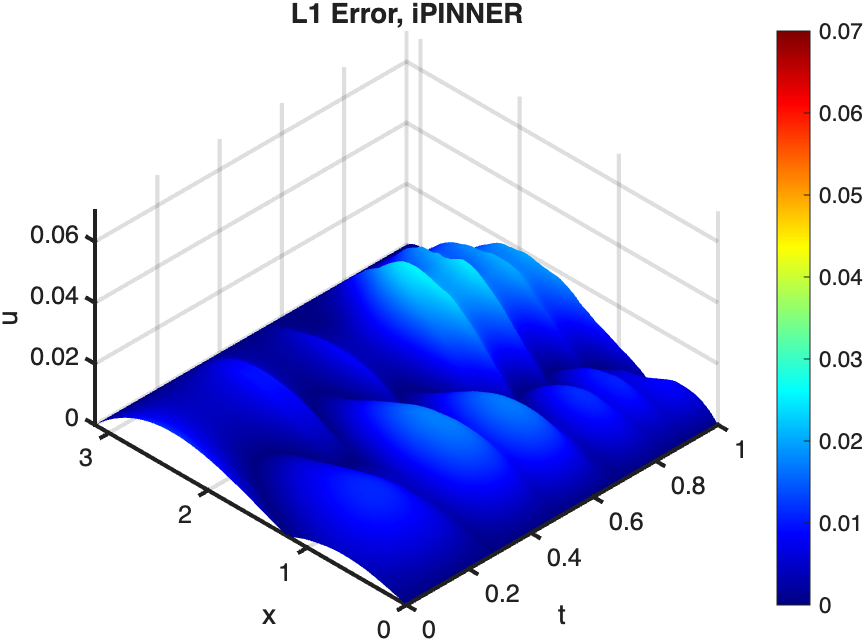}
        \caption{}
    \end{subfigure}
        \end{minipage}
    }
    \caption{
        Comparison of (a) the true solution and (b) the iPINNER solution as well as the $L_1$ errors of (c) the ADAM-PINN and (d) the iPINNER for  the time-fractional mixed diffusion-wave equations (TFMDWEs) in inverse problem.  Observation noise level is $20\%$.
    }
    \label{fig:fpde-inverse-20}
\end{figure}

\begin{figure}[H]
    \centering
    \resizebox{0.7\textwidth}{!}{  
    \begin{minipage}{\textwidth}
        \centering    
    \begin{subfigure}[b]{0.48\linewidth}
    \centering
        \includegraphics[width=\linewidth]{TFMDWEs_true_forward_noise20.png}
        \caption{}
    \end{subfigure}
    \hfill
    \begin{subfigure}[b]{0.48\linewidth}
    \centering
        \includegraphics[width=\linewidth]{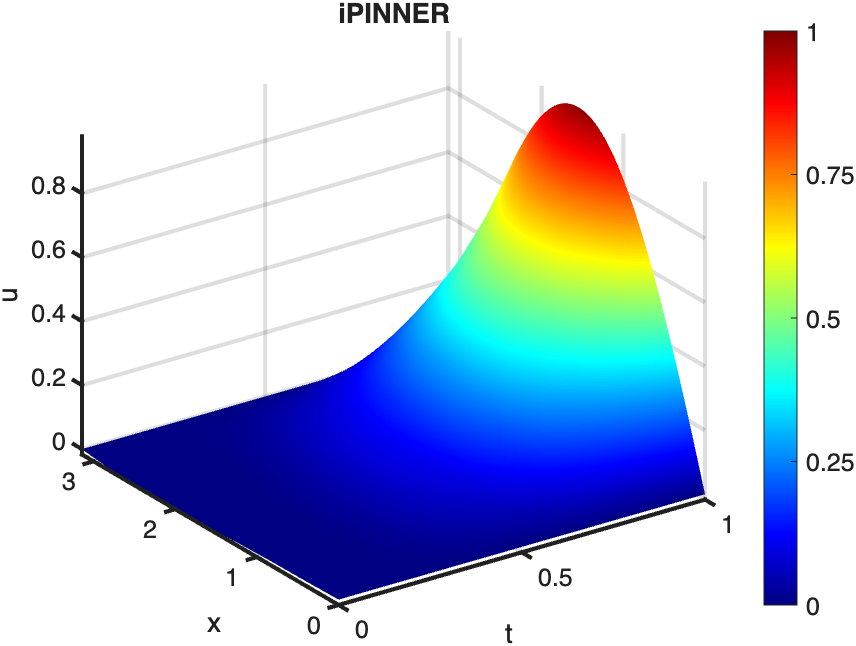}
        \caption{}
    \end{subfigure}
    \begin{subfigure}[b]{0.48\linewidth}
    \centering
        \includegraphics[width=\linewidth]{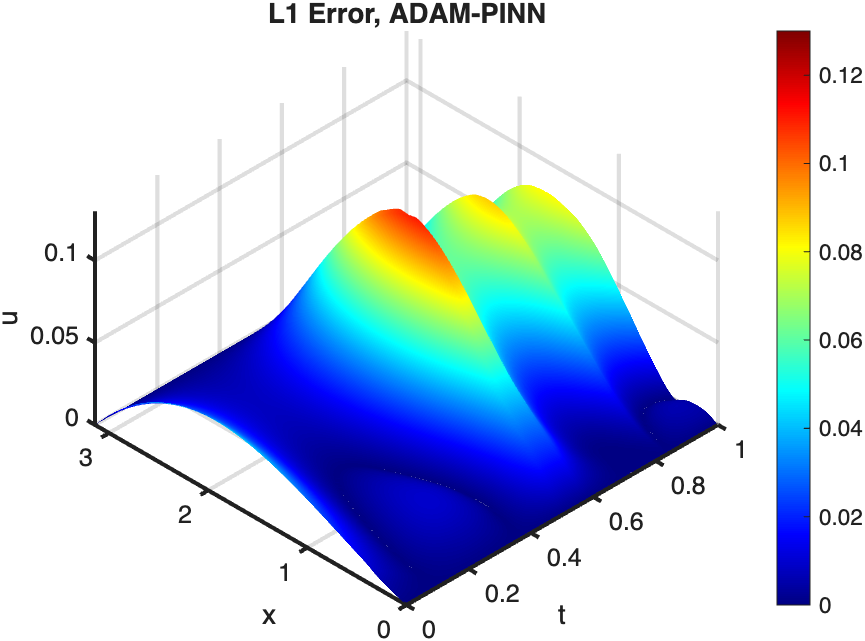}
        \caption{}
    \end{subfigure}
    \hfill
    \begin{subfigure}[b]{0.48\linewidth}
    \centering
        \includegraphics[width=\linewidth]{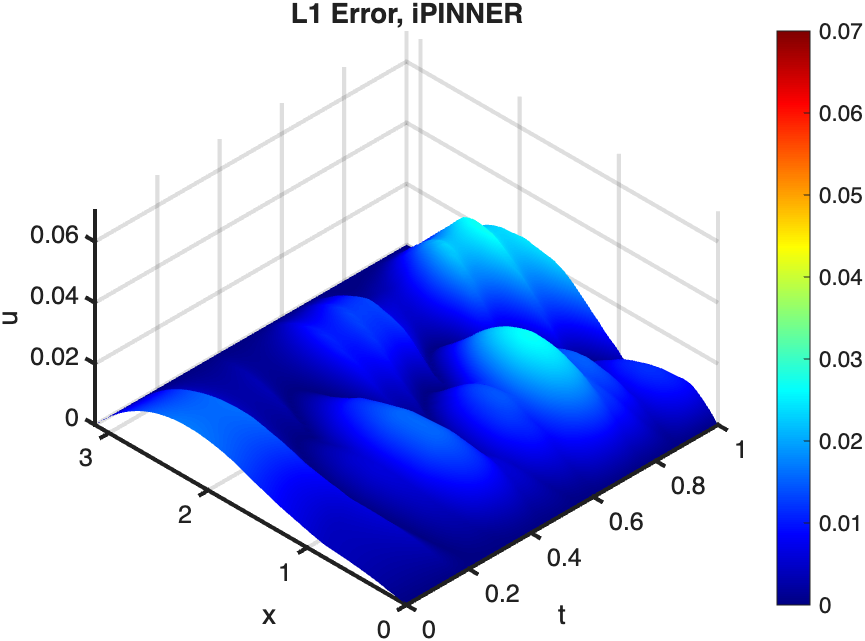}
        \caption{}
    \end{subfigure}
        \end{minipage}
    }
    \caption{
        Comparison of (a) the true solution and (b) the iPINNER solution as well as the $L_1$ errors of (c) the ADAM-PINN and (d) the iPINNER for  the time-fractional mixed diffusion-wave equations (TFMDWEs) in inverse problem. Observation noise level is $50\%$.
    }
    \label{fig:fpde-inverse-50}
\end{figure}

\begin{figure}[H]
    \centering
    \resizebox{0.7\textwidth}{!}{  
    \begin{minipage}{\textwidth}
        \centering
        \begin{subfigure}[b]{0.48\linewidth}
            \centering
            \includegraphics[width=\linewidth]{TFMDWEs_true_forward_noise20.png}
            \caption{}
        \end{subfigure}
        \hfill
        \begin{subfigure}[b]{0.48\linewidth}
            \centering
            \includegraphics[width=\linewidth]{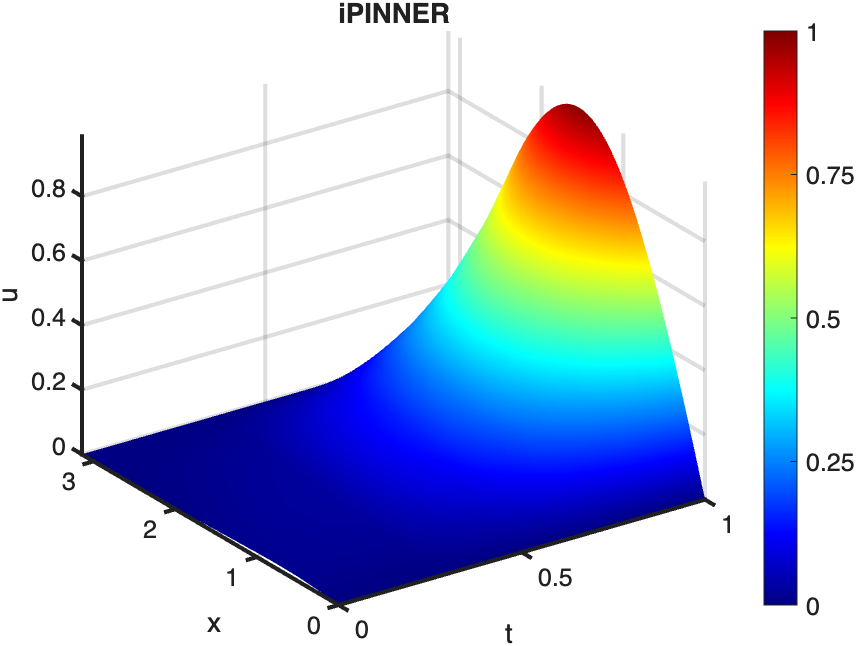}
            \caption{}
        \end{subfigure}

        \vspace{2mm}

        \begin{subfigure}[b]{0.48\linewidth}
            \centering
            \includegraphics[width=\linewidth]{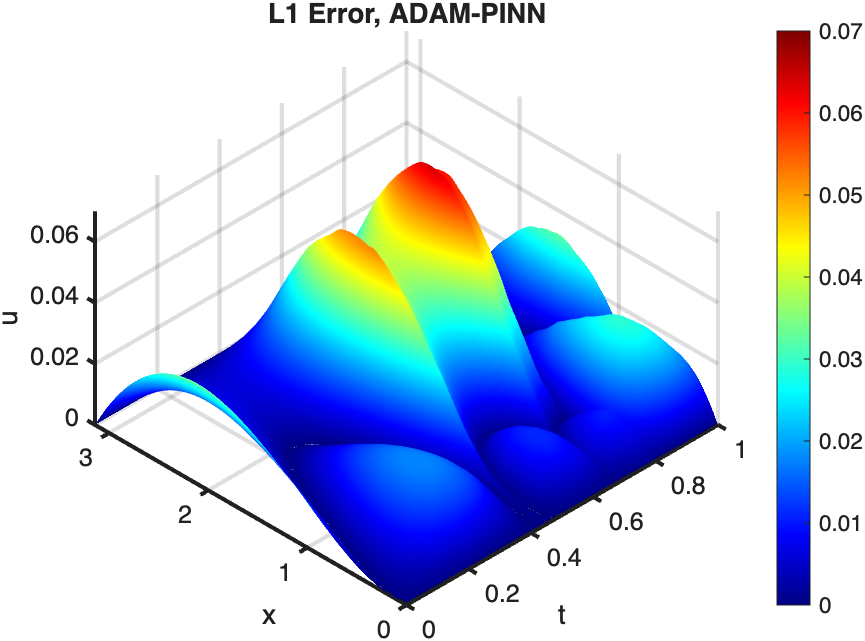}
            \caption{}
        \end{subfigure}
        \hfill
        \begin{subfigure}[b]{0.48\linewidth}
            \centering
            \includegraphics[width=\linewidth]{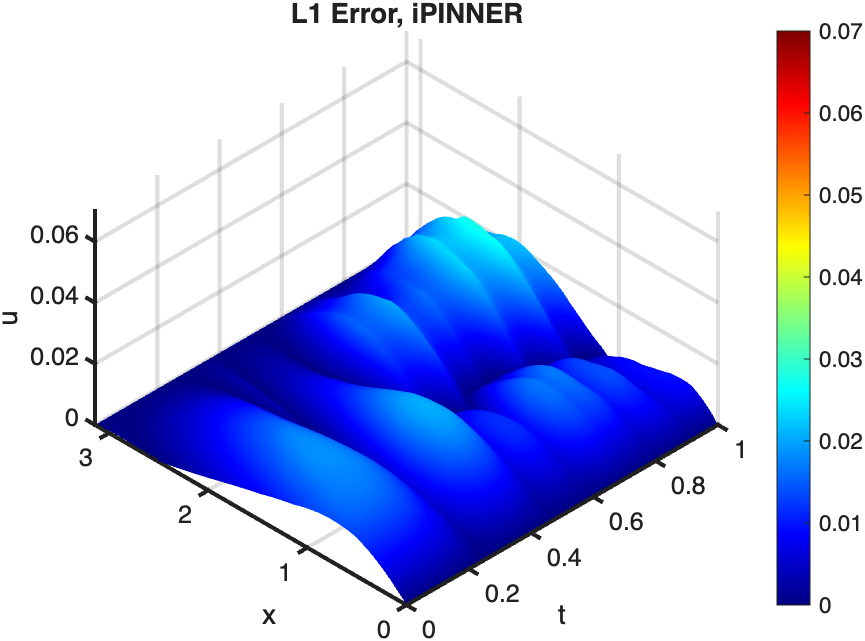}
            \caption{}
        \end{subfigure}
    \end{minipage}
    }

    \caption{
       Comparison of (a) the true solution and (b) the iPINNER solution as well as the $L_1$ errors of (c) the ADAM-PINN and (d) the iPINNER for  the time-fractional mixed diffusion-wave equations (TFMDWEs) in inverse problem.  Observation noise level is $80\%$.
    }
    \label{fig:fpde-inverse-80}
\end{figure}

\begin{figure}
    \centering
      \begin{subfigure}[b]{0.3\linewidth}
            \centering
    \includegraphics[width=\linewidth]{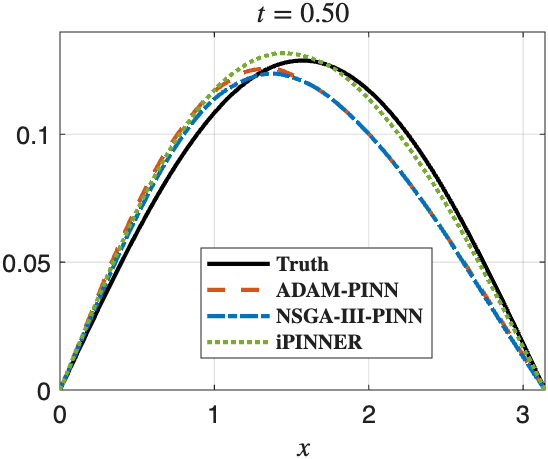}
    \includegraphics[width=.96\linewidth]{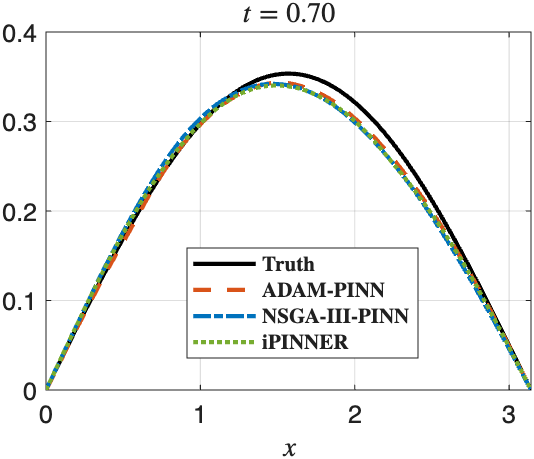}
                \caption{$20\%$ noise level }
        \end{subfigure}
              \begin{subfigure}[b]{0.3\linewidth}
            \centering
    \includegraphics[width=\linewidth]{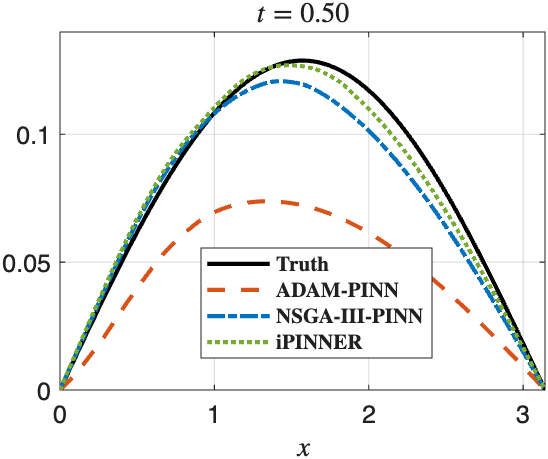}
    \includegraphics[width=.96\linewidth]{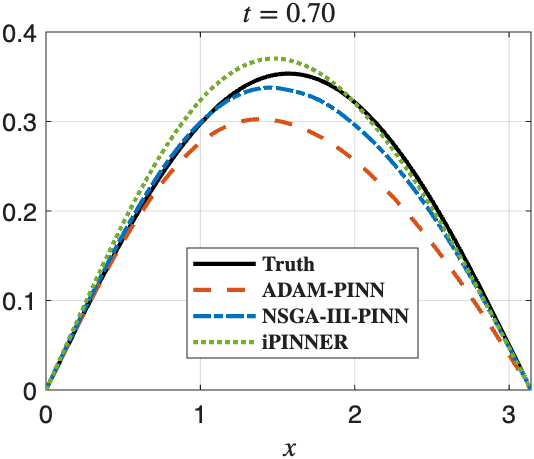}
                \caption{$50\%$ noise level }
        \end{subfigure}
              \begin{subfigure}[b]{0.3\linewidth}
            \centering
    \includegraphics[width=\linewidth]{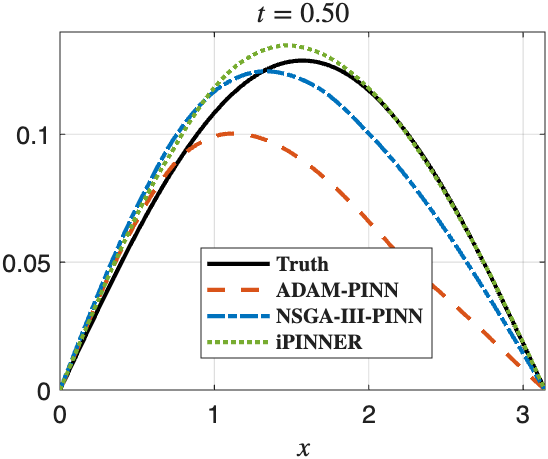}
    \includegraphics[width=.96\linewidth]{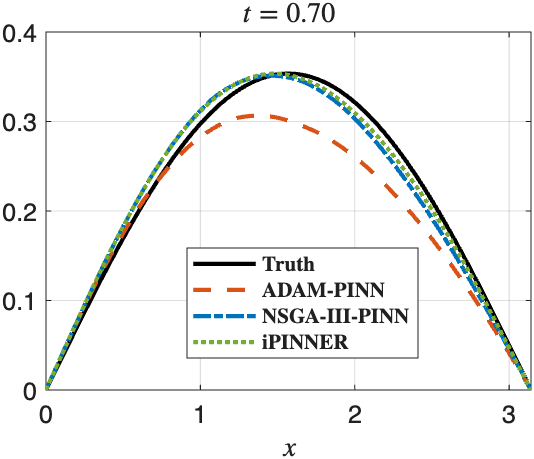}
                \caption{$80\%$ noise level }
        \end{subfigure}
    \caption{Comparison at different time instances (top: $t=0.5$, bottom: $t=0.7$) between the true solution (black) and inverse problem solutions obtained using ADAM-PINN (red), NSGA-III-PINN (blue), and iPINNER (green) for the time-fractional mixed diffusion-wave equations (TFMDWEs) under varying noise levels: Panel (a) — $20\%$ noise, Panel (b) — $50\%$ noise, and Panel (c) — $80\%$ noise. }
    \label{fig:fpde-curve-inverse}
\end{figure}

Table \ref{tab:fpde Inverse} shows the estimated $\alpha$ values of the TFMDWEs and their $L^1$ error  in the inverse problem setting for (1). ADAM-PINN, (2). NSGA-III-PINN, and (3). iPINNER with different noise level of observational data.
It is shown that in all cases, iPINNER is more accurate in estimating the missing parameters in TFMDWEs than ADAM-PINN and NSGA-III-PINN. 
Similar as in the forward problem setting, the NSGA-III-PINN shows more accurate estimations than ADAM-PINN, as the NSGA-III algorithm more effectively balances multiple objectives within the loss function, thereby partially mitigating the adverse effects of noisy observations even in the missing parameters. This is also reflected in the figures \ref{fig:fpde-inverse-20}--\ref{fig:fpde-inverse-80} as well as Figure \ref{fig:fpde-curve-inverse} which shows at different time instances (top: $t = 0.5$, bottom: $t = 1$), the true solution (black) and inverse problem
solutions obtained using ADAM-PINN (red), NSGA-III-PINN (blue), and iPINNER (green) for the time-fractional mixed diffusion-wave equations (TFMDWEs) with different noise levels. The results show that iPINNER (green) outperforms both ADAM-PINN (red) and NSGA-III-PINN (blue) in terms of estimating parameters and PDE solutions. 



\begin{table}[H]
    \centering
    \renewcommand{\arraystretch}{1.2} %
    \begin{tabular}{ccccc}
        \toprule
        Noise level &Model & $\alpha$ estimation &  $L^1$ error  \\

        \midrule
        \multirow{3}{*}{20\%} &  ADAM-PINN    & 0.5556 &0.0556    \\
                              &  NSGA-III-PINN    & 0.5440  & 0.0440   \\
                              &   \textbf{iPINNER} &\textbf{0.5219}   & \textbf{0.0219}  \\
        \midrule
        \multirow{3}{*}{50\%} & ADAM-PINN     &0.5760   & 0.0760  \\
                              & NSGA-III-PINN   &0.5657   &  0.0657 \\
                              &  \textbf{iPINNER} &\textbf{0.5450 }  & \textbf{0.0450} \\
        \midrule                      
        \multirow{3}{*}{80\%} & ADAM-PINN    &0.6208  &  0.1208  \\
                              & NSGA-III-PINN     &0.6050   & 0.1050   \\
                              & \textbf{iPINNER} & \textbf{0.5713}  & \textbf{0.0713}\\
        \bottomrule
    \end{tabular}
    \caption{Comparison of PINN with Adam optimizer (ADAM-PINN), PINN with NSGA-III optimizer (NSGA-III-PINN) and the proposed integrated method (iPINNER) for estimating the fractional term $\alpha$ in the TFMDWEs. For reference, the true value of the fractional term is $\alpha=0.5$.}
    \label{tab:fpde Inverse}
\end{table}

\subsection{Two-Dimensional Heat Equation}
\label{sec:2d-heat}
We next consider the classical two-dimensional heat equation on the unit square with homogeneous Dirichlet boundaries:
\begin{align}
    \label{eq:2d-heat}
&\partial_t u(x,y,t) - \kappa \, \Delta u(x,y,t) \;=\; f(x,y,t),
\qquad (x,y)\in\Omega\equiv [0,1]^2,\; t\in[0,1],\\
&u|_{\partial\Omega}=0, 
\\
&u(x,y,0)=u_0(x,y).
\end{align}
To enable quantitative comparison, we consider the analytical solution
\begin{equation}
u(x,y,t) = \sin(\pi x)\sin(\pi y)\, \mathrm{e}^{-2\pi^2\kappa t},
\end{equation}
which corresponds to non forcing case, i.e., $f \equiv 0$ and the initial condition $u_0(x,y) = \sin(\pi x)\sin(\pi y)$. 

\paragraph{Neural network architecture} 
In order to make a fair comparison, with different optimizers, the PINN employs a consistent deep neural network architecture comprising 6 layers with 64 neurons in each layer.
During the training phase, models corresponding to each optimizer were collected until the training loss converged below a prescribed threshold $\epsilon$.
Specifically, the ADAM-PINN requires $30000$ epochs to achieve convergence. The proposed iPINNER requires of $5$ generations with $5000$ epochs to ensure the convergence of training loss.

\paragraph{Training and Testing Data} The training data for the 2D Heat equation consist of three main components: Initial Condition (IC) points, Boundary Condition (BC) points, and Collocation Points. The IC points are sampled in space at the initial time $t = 0$, while the BC points are sampled in time along the spatial boundaries $\partial\Omega$. The collocation points are randomly selected from the interior of the spatiotemporal domain, with 100 points used in this study. The testing data are sampled over the domain $\Omega \times [0,T]$ with $\Omega = [0,1]$.  The spatial mesh size is of $h=\Delta x=\Delta y = 1/10$ and a temporal step size is $\Delta t = 0.01$.

\paragraph{Error criteria} We evaluate models using the $L_2$ relative error between the benchmark solutions and the predictions of different models.

In the forward problem setting,
we samples the observational data at $N_{\text{obs}}$ scattered space–time locations, which are then added with zero-mean Gaussian noise with standard deviation equal to a prescribed fraction $\eta \in \{20\%, 50\%\}$ of $\mathrm{std}(u)$ evaluated at those points (consistent with Sections~\ref{sec:burgers}–\ref{sec:frac}). 

\begin{figure}[H]
    \centering
    \includegraphics[width=1.0\linewidth]{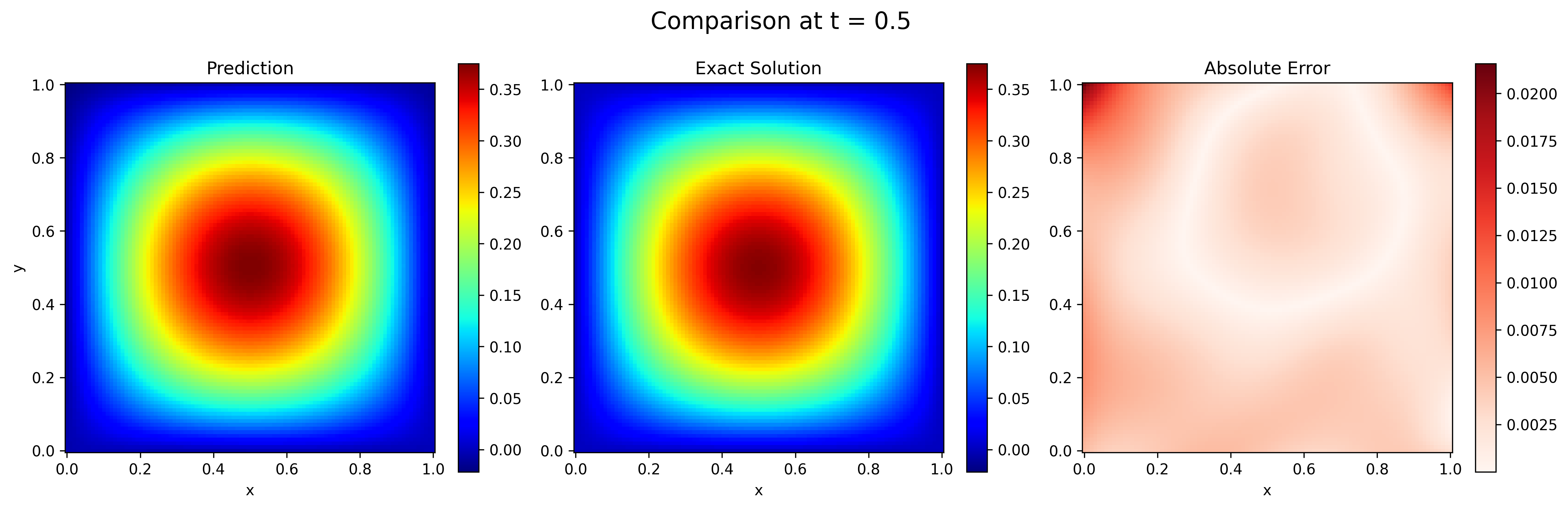}
    \includegraphics[width=1.0\linewidth]{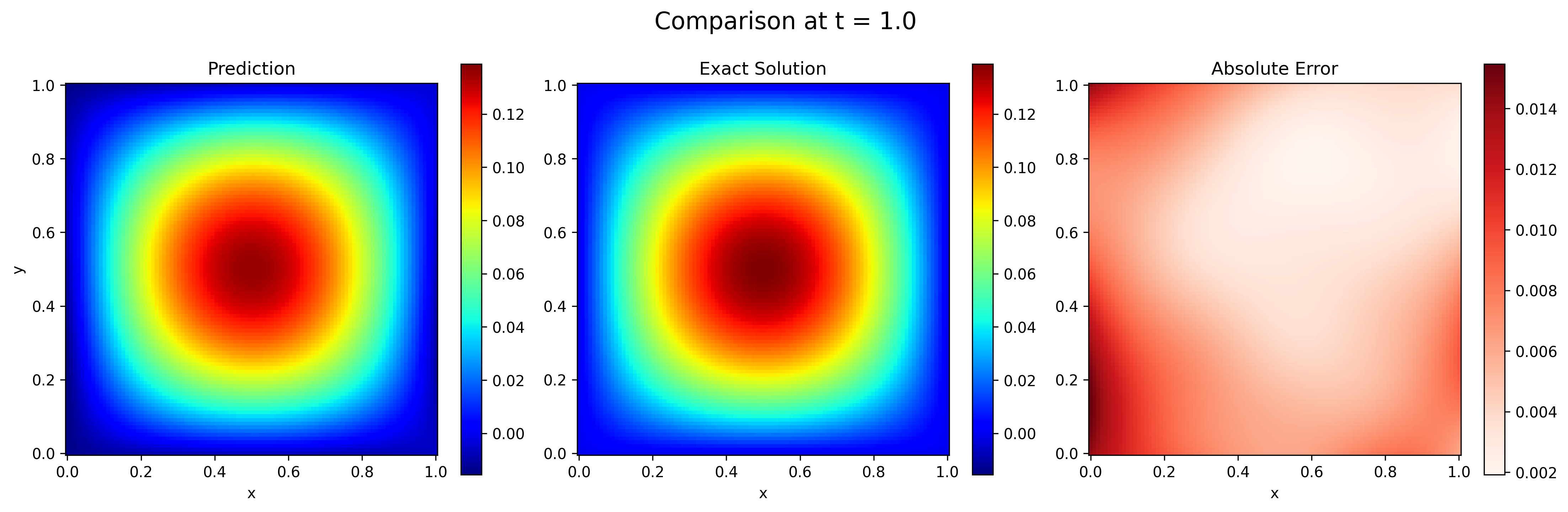}
    \caption{iPINNER prediction of 2D Heat equation with 20\% noise data.}
    \label{fig:heat-1}
\end{figure}

\begin{figure}[H]
    \centering
    \includegraphics[width=1.0\linewidth]{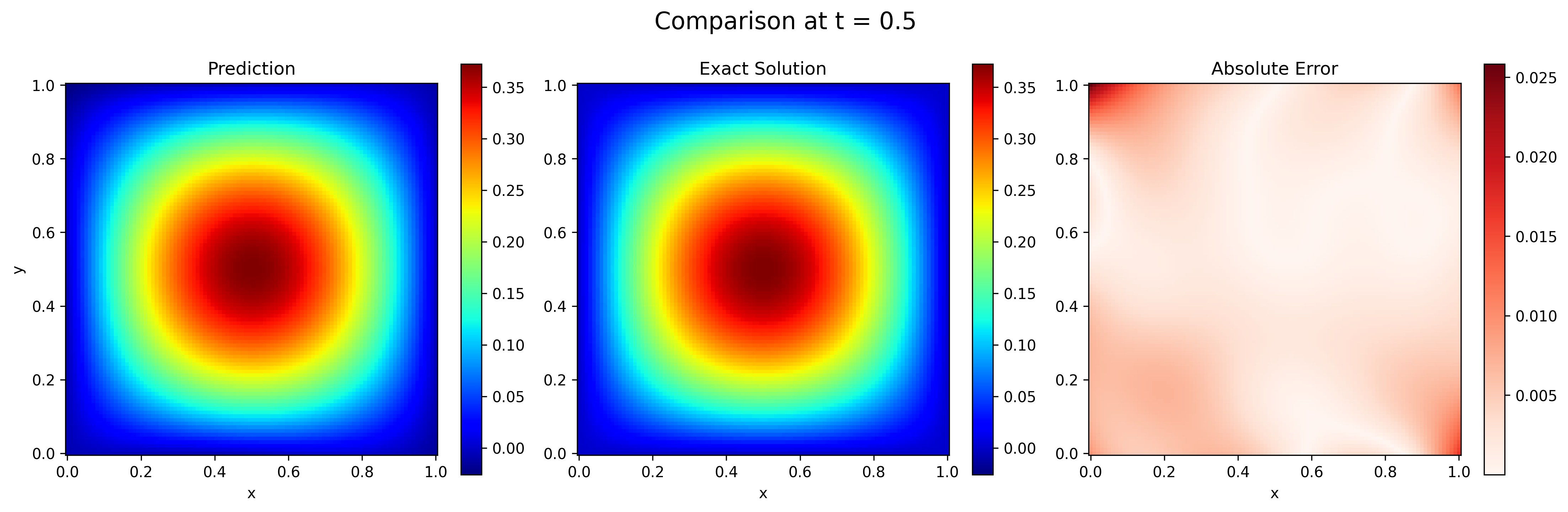}
    \includegraphics[width=1.0\linewidth]{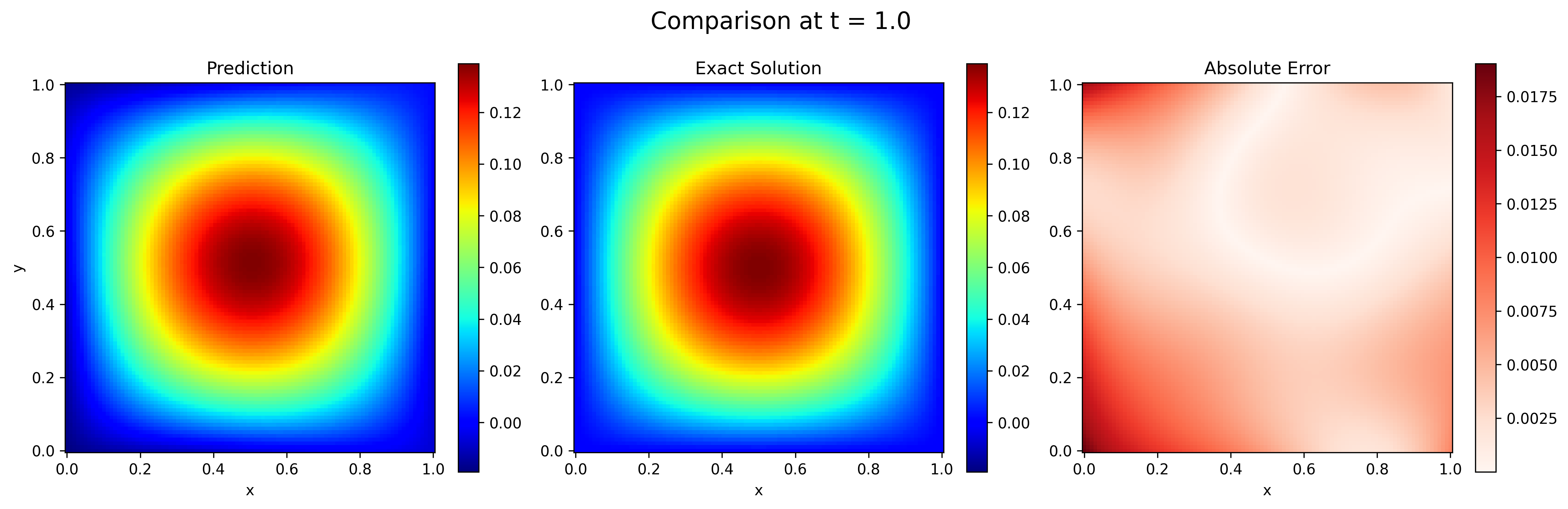}
    \caption{iPINNER prediction of 2D Heat equation with 50\% noise data.}
    \label{fig:heat-2}
\end{figure}

Figures~\ref{fig:heat-1}–\ref{fig:heat-2} show the forward problem solutions using iPINNER (left) and the corresponding true solutions (middle) of the 2D heat equation at two different time instances (top: $t = 0.5$, bottom: $t = 1$) under different noise levels. The right panels show the corresponding $L^1$ errors between the predicted and true solutions. iPINNER shows higher accuracy than ADAM-PINN, particularly under higher noise levels. With 50\% data noise, the iPINNER has 0.0180 $L_2$ relative error whereas ADAM-PINN has 0.0285 $L_2$ relative error shown in Table~\ref{tab:walltime_comparisonHeat}

\begin{table}[htbp]
\centering
\caption{Wall time of training PINN and iPINNER with 50\% noise data.}
\label{tab:walltime_comparisonHeat}
\begin{tabular}{lcc}
\toprule
\textbf{Method} & \textbf{Wall Time (s)} & \textbf{$L_2$ relative error} \\
\midrule
Standard PINN    & 256  &0.0285  \\
\textbf{iPINNER} & \textbf{430}& \textbf{0.0180} \\
\bottomrule
\end{tabular}
\end{table}

\section{Conclusion and Future Work \label{sec:conclusion}}

In this paper, we introduce a novel iPINNER framework that integrates the \textit{physics informed neural network} (PINN) with ensemble Kalman filter (EnKF) with the NSGA-III multi-objective optimizer. This framework can be used to address both forward problem and inverse problem with only limited noisy data in the context of partial differential equations (PDE). 
The iPINNER framework has several advantages: (i) The framework utilizes the multi-objective optimizer NSGA-III to find an optimal cluster of PINNs, and (ii) The uncertainty within this cluster, inherent from the NSGA-III optimization, can be integrated with observational noisy data using the ensemble Kalman filter. This process can then be iteratively applied to update the data loss during PINN training. 
(iii) The framework can be applied in both the forward and inverse problems in solving PDEs.

Building on the iPINNER framework's effectiveness in solving forward and inverse problems with noisy data, several future directions necessitate further investigation.
First, in the inverse problem setting, the unknown parameters are treated as trainable variables within the neural networks and are implicitly embedded in the PDE residual loss of the PINN. To quantify the uncertainty of the inferred parameters, these parameters together with the independent variables can be used as inputs to the neural network, allowing it to be trained across a wide range of physical parameters \cite{harlim2021machine,gottwald2021combining}.
Second, a Bayesian PINN framework for both inverse and forward models, as proposed in \cite{yang2021b}, can be integrated with the current framework. This integration could potentially improve the performance of the model in the presence of highly noisy data, as the Bayesian approach naturally quantifies the uncertainties arising from scattered noisy data.
Third, the iPINNER framework can be extended to a continual learning framework, allowing it to adapt and improve over time \cite{purvine2017comparative,howard2024multifidelity,roy2024exact}. In this approach,  PINNs optimized with the NSGA-III algorithm can be incrementally updated with new observational data. This continual learning process allows the model to refine its predictions, improve accuracy, and maintain robustness when additional data becomes available. Moreover, it facilitates the dynamic update of systems and uncertainties, making it particularly valuable for real-time prediction and long time simulation of complex physical systems \cite{ghunaim2023real,shaheen2022continual,yang2012adaptive}. 

\section*{Acknowledgment}
Guang Lin acknowledges the National Science Foundation under grants DMS-2053746, DMS-2134209, ECCS-2328241, CBET-2347401, and OAC-2311848. The U.S. Department of Energy also supports this work through the Office of Science Advanced Scientific Computing Research program (DE-SC0023161) and the Office of Fusion Energy Sciences (DE-SC0024583).

\bibliographystyle{abbrv}
\bibliography{ref,ref_pinn}

\appendix
\section{Mathematical Formulation of Ensemble Kalman Filter (EnKF)\label{app:enkf}}
Consistent with the notation in Section~\ref{ss:enkf}, let the forecast ensemble at time $t$ be denoted as follows
\begin{align}
X_{t|t-1} = \bigl[x_{t|t-1}^{(1)}, , x_{t|t-1}^{(2)}, , \dots, , x_{t|t-1}^{(N)}\bigr] \in \mathbb{R}^{d \times N},
\end{align}
where $N$ is the ensemble size and $d$ denotes the dimension of the state variable. Each column $x_{t|t-1}^{(i)}$ represents one realization of the state vector in forecast models.
Then, the prior mean $\bar{x}_{t|t-1}$ and covariance $R_{t|t-1}$ (model forecasts) are then estimated by the following \cite{evensen2003ensemble}
\begin{align}
\bar{x}_{t|t-1} &= \frac{1}{N}\sum{i=1}^N x_{t|t-1}^{(i)}, 
\\
R_{t|t-1} &= \frac{1}{N-1}\sum_{i=1}^N \bigl(x_{t|t-1}^{(i)} - \bar{x}_{t|t-1}\bigr)\bigl(x_{t|t-1}^{(i)} - \bar{x}_{t|t-1}\bigr)^\top,
\end{align}
which provide Monte Carlo approximations to the prior mean and covariance. 
Then, given an observation $y_t$, the ensemble members can be updated through Bayesian formula and yields the following:
\begin{align}
x_t^{(i)} = x_{t|t-1}^{(i)} + K_t \bigl(y_t^{(i)} - H x_{t|t-1}^{(i)}\bigr), \quad i=1,\dots,N,
\end{align}
where $y_t^{(i)} = y_t + \epsilon_t^{(i)}$ are observations with observation error $\epsilon_t^{(i)} \sim \mathcal{N}(0,R)$. The ensemble Kalman gain can be derived as follows:
\begin{align}
K_t = P_{t|t-1} H^\top \bigl(H P_{t|t-1} H^\top + R\bigr)^{-1}.
\end{align}

\end{document}